\documentclass[journal]{IEEEtran}
\usepackage{amsmath,amsfonts}
\usepackage{algorithmic}
\usepackage{algorithm}
\usepackage{array}
\usepackage[caption=false,font=normalsize,labelfont=sf,textfont=sf]{subfig}
\usepackage{textcomp}
\usepackage{stfloats}
\usepackage{url}
\usepackage{verbatim}
\usepackage{graphicx}
\usepackage{cite}
\usepackage{bm}
\usepackage{nicefrac}
\usepackage[table,xcdraw]{xcolor}
\usepackage{hyperref}
\usepackage{booktabs}
\begin{document}
\title{Machines Serve Human: A Novel Variable Human-\\machine Collaborative Compression Framework}
\author{Zifu Zhang,~\IEEEmembership{Student Member,~IEEE,} Shengxi Li$^{*}$,~\IEEEmembership{Member,~IEEE,} Xiancheng Sun, \IEEEmembership{Student Member,~IEEE,} \\Mai Xu,~\IEEEmembership{Senior Member,~IEEE,} Zhengyuan Liu, Jingyuan Xia,~\IEEEmembership{Member,~IEEE}
\thanks{Zifu Zhang, Shengxi Li (Corresponding author), Xiancheng Sun, Mai Xu and Zhengyuan Liu are with the School of Electronic and Information Engineering, Beihang University, Beijing 100191, China (Email: \{ZifuZhang; LiShengxi; xianchengsun; MaiXu; zy2302223\}@buaa.edu.cn).  Shengxi Li is also with State Key Laboratory of Virtual Reality Technology and Systems, Beihang University. Jingyuan Xia is with the Department of Electronic Science and Technology, National University of Defense Technology, Changsha 410073, China (e-mail: j.xia10@nudt.edu.cn). This work was supported by NSFC under Grants 62450131, 62206011 and 62231002, and Beijing Natural Science Foundation under Grant L223021.}
 }

\markboth{Journal of \LaTeX\ Class Files,~Vol.~14, No.~8, August~2021}%
{Shell \MakeLowercase{\textit{et al.}}: A Sample Article Using IEEEtran.cls for IEEE Journals}


\maketitle

\begin{abstract}
Human-machine collaborative compression has been receiving increasing research efforts for reducing image/video data, serving as the basis for both human perception and machine intelligence. Existing collaborative methods are dominantly built upon the \textit{de facto} human-vision compression pipeline, witnessing deficiency on complexity and bit-rates when aggregating the machine-vision compression. Indeed, machine vision solely focuses on the core regions within the image/video, requiring much less information compared with the compressed information for human vision. In this paper, we thus set out the first successful attempt by a novel collaborative compression method based on the machine-vision-oriented compression, instead of human-vision pipeline. In other words, machine vision serves as the basis for human vision within collaborative compression. A plug-and-play variable bit-rate strategy is also developed for machine vision tasks. Then, we propose to progressively aggregate the semantics from the machine-vision compression, whilst seamlessly tailing the diffusion prior to restore high-fidelity details for human vision, thus named as diffusion-prior based feature compression for human and machine visions (Diff-FCHM). Experimental results verify the consistently superior performances of our Diff-FCHM, on both machine-vision and human-vision compression with remarkable margins. Our code will be released upon acceptance.
\end{abstract}


\section{Introduction}

\IEEEPARstart{T}{he} rapid advancement of multimedia services and applications has led to an exponential increase in the volume of image and video data, posing significant challenges against transmission bandwidth and storage requirements. Efficient image/video compression is thus imperative to reduce the large data volume whilst improving subjective quality for human vision. Correspondingly, the past decades have witnessed successive advanced standards including high efficiency video coding (HEVC) \cite{sullivan2012overview} and versatile video coding (VVC) \cite{bross2021overview}, together with recent end-to-end learned compression methods in parallel \cite{balle2018variational, zhang2025continuous, lu2019dvc, cheng2020learned}. On the other hand, the BigData area has revolutionized the way the intelligent machines approach the world \cite{han2025generating}, in which compression for machine vision tasks rather than human vision, is also extensively investigated; those include video coding for machines (VCM) \cite{kang2022vcm} and feature coding for machines (FCM) \cite{CTCstandard}  as the emerging standards by moving picture experts group (MPEG). More importantly, to achieve the universal compression, the efficient human-machine collaborative compression has been playing as the central role catering for both human and machine vision, which has been intensively discussed most recently \cite{he2024learned, li2024humanJESTA, li2024human}. 

{The compressed images for human vision are able to operate as a satisfied input towards existing machine vision pipelines, thus catering for machine vision tasks. In other words, compression for human vision constitutes a natural basis for machine-vision-oriented compression. Thus, existing collaborative methods are almost built upon human-vision-based compression frameworks, in which follow-up optimisation was proposed regarding machine vision tasks, including task-specific prompt tuning \cite{chen2023transtic}, frequency component integration \cite{li2024image} and hierarchical scalable coding \cite{liu2023icmh}. However, for human vision, retaining high-quality compression typically requires extra bit-rates to restore image details, which are typically redundant for machine-vision-oriented compression. Collaborative compression upon human vision thus exhibits deficiency on both computational complexity and bit-rates, compared to the compression methods purely designed for machine vision \cite{zhang2024hybrid, chen2023residual}. }

\begin{figure*}[htbp]
   \begin{center}
   \includegraphics[width=0.9\linewidth]{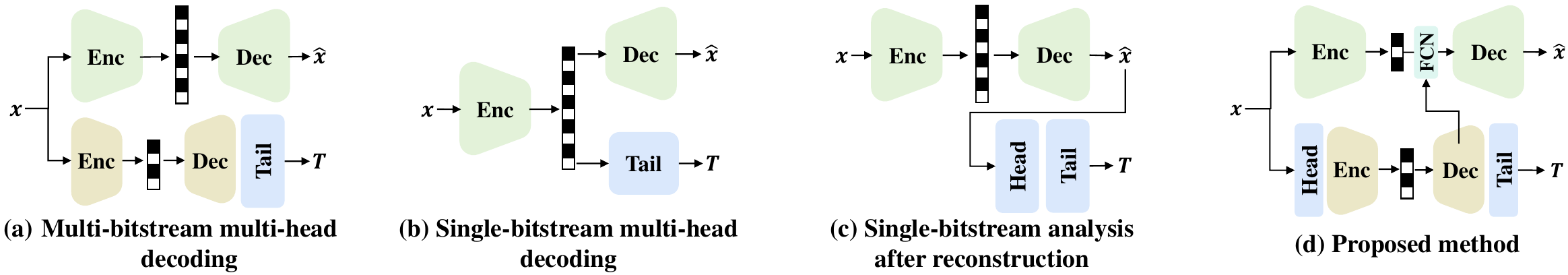}
   \end{center}
      \caption{{Illustration of existing human-machine collaborative compression paradigms, against our framework. Enc and Dec denote the encoder and decoder, Head and Tail refer to the initial and final blocks of the machine-vision network, FCN represents fusion control network, $\bm{x}$, $\hat{\bm{x}}$, and $\bm{T}$ denote the original image, reconstructed image, and downstream task results, respectively. Please note that $\hat{\bm{x}}$, and $\bm{T}$ correspond to compression for human vision and machine vision, respectively.}}
   \label{fig:fig1}
\end{figure*}

{Indeed, the core visual cues emphasized in machine vision tasks, including detected objects and segmented masks, align closely with the region-of-interest regions for human perception. This thus motivates us to design collaborative compression via new principles, i.e., \textit{machine vision serves as the basis for human vision}. We then establish the collaborative method based on the machine-vision compression framework, instead of the human-vision compression pipelines, with significant potential of improving compression efficiency for both human and machine vision. Despite this, recent works \cite{li2024human, guo2023toward} reveal that collaborative compression based on machine-vision compressed features still remains highly challenging, because deep-learning-based representations, oftentimes capturing rich semantic information, struggle to preserve low-level details such as textures, edges and pixel-level colour, resulting in visual artifacts in the decoded images for human vision. This essentially leads to the semantic-fidelity trade-off inherited from the machine-vision features.}

{In this paper, we thus propose a new human-machine collaborative compression framework, on the basis of machine-vision compressed features, and leverage the human-vision priors from deep generative models \cite{ho2020denoising} that are capable of generating high-quality details given the compressed semantic features for machine vision tasks. This way, the bit-rate can be maximally saved whilst the reconstruction simultaneously retains best accuracy for machine vision and high fidelity for human vision. To the best of our knowledge, the proposed method is the first successful attempt to achieve collaborate compression based on machine-vision-based features, which paves a new way of designing human-machine collaborative compression. We thus name our method as Diff-FCMH, the abbreviation of diffusion-prior based feature compression for machine and human visions.} 
Experimental results verify the superior performances of our method, on subjective quality and semantic consistency for human visions, together with the
state-of-the-art accuracy on machine vision tasks.  Our main contributions are mainly three-fold:
\begin{itemize}
\item We develop a simple yet effective variable-rate compression strategy with implicit normalisation and denormalisation for machine vision, which is able to achieve variable-rate compression without requiring any additional training procedures.
\item We propose to efficiently aggregate the semantics from machine-vision-oriented compression in an autoregressive manner, which enhances semantic alignment with natural image distributions for human vision. 
\item We propose a new collaborative compression paradigm from the machine-vision perspective, via seamlessly incorporate the diffusion prior for human vision. The superiority of our method is remarkable, resulting in consistent gains for both machine and human vision tasks.
\end{itemize}

\section{Related Works}

\subsection{Compression for Human-Machine Vision}

Since human vision and machine vision are both important for reducing image data, human-machine collaborative compression has received increasing research efforts, which can be mainly categorized into 3 categories \cite{li2024human} with dual branches, as also shown in Fig. \ref{fig:fig1}, namely, multi-bitstream multi-head decoding (MBMD) \cite{duan2015overview, cao2023slimmable, liu2022improving}, single-bitstream multi-head decoding (SBMD) \cite{ bai2022towards, hu2020towards} and single-bitstream analysis after reconstruction (SBAR) \cite{chen2023transtic, li2024image, guo2023toward} strategies. More specifically, the MBMD strategy adopts separate encoders for human and machine vision to independently compress image content and task-specific features. This results in multiple task feature bitstreams and single image reconstruction bitstream, which are subsequently decoded by distinct decoders tailored for each task. 
Compared to MBMD, the SBMD strategy employs a single encoder to generate a unified bitstream, which is then decoded by separate decoders for image reconstruction and machine vision tasks. Several works utilize subsets of the human-vision bitstream to perform machine vision tasks, enabling scalable hierarchical decoding \cite{liu2023icmh, choi2022scalable} but degrading task-specific performance. Moreover, the SBAR methods refer to fine-tuning human-vision compression networks for machine vision tasks, by incorporating strategies such as prompt tuning \cite{chen2023transtic} and spatial-frequency modulation \cite{li2024image}. However, the overall framework is still based on the human-vision compression pipeline, with minor adjustments regarding machine vision tasks. This essentially leads to redundancy across different compression tasks, since bit-rates from human-vision compression are typically much larger than those for machine vision tasks.

Recent years, generative models have undergone remarkable advancements \cite{liu2025causal}, with generative adversarial networks (GANs)  \cite{goodfellow2020generative} and diffusion models \cite{song2020denoising} serving as two core driving force in this field. The advancements opened up new possibilities for achieving ultra-low bitrate compression optimized for human visual perception. Existing works \cite{rippel2017real, mentzer2020high, agustsson2023multi} have leveraged GAN loss to significantly enhance the perceptual quality of compressed images. With the rapid progress of large-scale text-to-image diffusion models, recent studies \cite{lei2023text+sketch, li2024misc, careil2023towards} explore the rich semantics embedded in pretrained diffusion models for generative compression at extremely low bitrates. However, the above methods are mainly designed for human visual perception, with limited exploration in human–machine collaborative compression. Several GAN-based approaches \cite{zhang2025HSC, mao2023scalable} have investigated joint compression for human and machine vision on face datasets, whereas broader applications to machine vision tasks such as detection and segmentation remain largely unexplored \cite{han2024large}. 

\subsection{Feature Compression for Machine Vision}

For machine-vision-oriented compression, feature coding for machiens (FCM) is developed for compressing features from vision tasks, oftentimes achieving significantly better bit-rate savings than traditional video coding for machines (VCM) \cite{datta2022low, wu2024scalable}. VCM is based on an image compression framework, indirectly enhancing the performance of machine vision tasks by optimizing the rate-distortion objective at the pixel level. However, the advantage of FCM lies in its focus on extracting and compressing task-relevant features, rather than preserving the pixel-level fidelity required for human visual perception \cite{zhangall}.  Correspondingly, the recent progress of MPEG calls for efficient methods regarding FCM and the typical backbone for machine vision tasks is the same as the setting for VCM  \cite{CTTCstandard}. Several proposals focus on reducing the redundancy across scales of features, using learned multi-scale feature compression network \cite{kim2023end}, asymmetrical feature coding \cite{zhang2024afc} and hybrid single input and multiple output structure \cite{zhang2024hybrid}. These approaches aim to efficiently represent multi-scale semantic features while maintaining the discriminative power necessary for downstream machine vision tasks such as object detection and instance segmentation. However, the compressed features are tailored specifically for certain machine vision tasks on specific networks like Faster R-CNN \cite{ren2015faster} or Mask R-CNN \cite{he2017mask}, making it infeasible to reconstruct images that meet human visual requirements \cite{cao2023slimmable, li2024human}. Thus, although FCM methods can achieve significant bit-rate saving, their intrinsic design to machine vision tasks prevents it from recovering details within images for human vision \cite{yan2021sssic}.

\subsection{Variable Bit-rates Image Compression}

Deep learning-based compression models typically support a single compression bitrate per model, which often requires substantial training resources. To address this limitation, various methods have been proposed to achieve variable-rate compression using a single learning-based model. In an early approach, Choi et al. \cite{choi2019variable} introduced a conditional autoencoder with an adjustable quantization bin size, extending fixed-rate models to a narrow range of continuous bitrates without significant performance degradation. Subsequent studies \cite{cui2021asymmetric, akbari2021learned, duan2023qarv} have followed the amplitude modulation paradigm, which defines scaling coefficients or subnetworks to adjust the amplitude of latent representations, thereby indirectly controlling quantization error and achieving variable bitrates. For instance, Tong et al. \cite{tong2023qvrf} proposed a quantization-error-aware variable-rate framework (QVRF) that employs a univariate quantization regulator to realize wide-range variable rates within a single model. More recently, Tu et al. \cite{tu2025msinn} designed a lightweight multi-scale invertible neural network that bijectively maps the input image into multi-scale latent representations, achieving superior compression performance compared to VVC across a very wide range of bitrates with a single model. However, existing methods still involve complex training procedures, often requiring staged sampling of different quantization parameters. Moreover, previous studies have primarily focused on compression strategies tailored for human visual perception, and have yet to explore variable-rate compression strategies specifically designed for machine vision tasks.

\begin{figure*}[htbp]
   \begin{center}
   \includegraphics[width=1\linewidth]{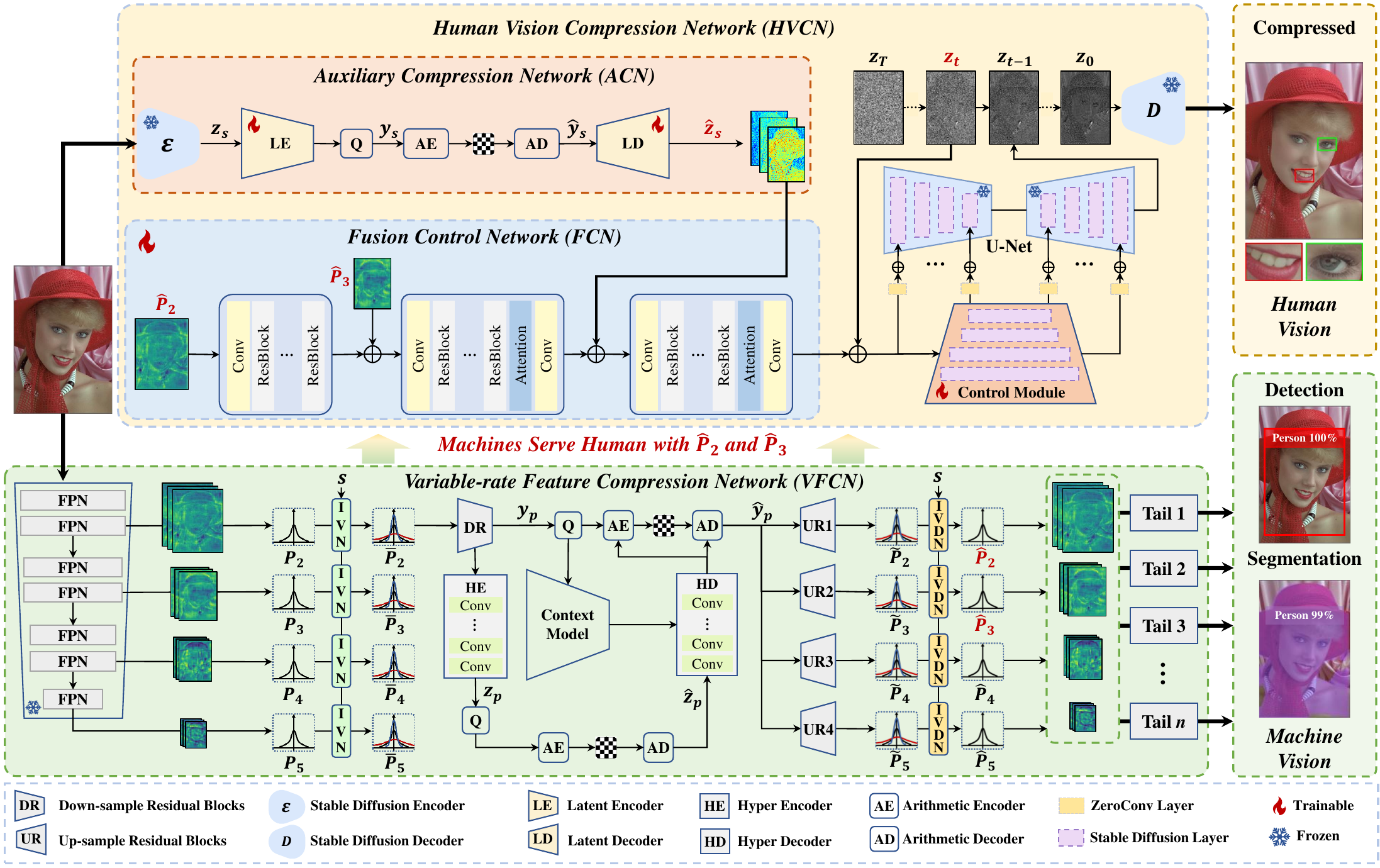}
   \end{center}
      \caption{{The overall architecture of our Diff-FCMH method, which first obtains machine-vision features from the task head. The machine-vision features are then compressed via our variable-rate feature compression network (VFCN), through implicit variable normalisation (IVN) and  de-normalisation (IVDN) layer, enabling variable downstream task performance through the remaining task tail networks. The compressed machine-vision features then operate as the basis for the human vision compression network (HVCN), with the goal of achieving high-fidelity compression for human vision. This is achieved by our newly proposed fusion control network (FCN) module and auxiliary compression network (ACN) module, which progressively aggregate the diffusion prior and the noisy latent with the semantics from machine-vision features.} }
   \label{fig:framework}
\end{figure*}

\section{Proposed Method}

\subsection{Motivation of Overall Architecture}

{The overall architecture of our network is illustrated in Fig. \ref{fig:framework}, which mainly consists of variable-rate feature compression network (VFCN) and human vision compression network (HVCN). More specifically, in our VFCN as regularized by the MPEG VCM/FCM group \cite{CTTCstandard}, the multi-scale features $\{{\bm{P}}_i\}_{i=2}^5$ are first extracted by the Faster R-CNN (or Mask R-CNN) backbone for object detection (or instance segmentation), as the task head of our Diff-FCMH method. To enable variable bit-rates in our machine vision compression network, we introduce an implicit  variable normalisation (IVN) layer that tailors the input distribution before the encoder, instead of the output distribution after the encoder, so as to achieve quantisation. Rescaling at input level is particularly effective for machine vision tasks when controlling bitrates, by implicitly highlighting task-oriented region of interest within input features.  The IVN layer uniformly scales features using a global scaling factor, producing normalized feature representations $\{{\bm{\bar{P}_i}}\}_{i=2}^5$. Moreover, given the pyramid architecture within vision tasks, the features across different scales exhibit large redundancy. We thus only compress the largest-scale feature, i.e., $\bar{\bm{P}}_2$, via a hyper-prior network with an auto-regressive context model, whilst simultaneous reconstructing features of all scales $\{{\bm{\tilde{P}}}_i\}_{i=2}^5$, so as to significantly save the bit-rates for machine-vision compression. Correspondingly, an  implicit variable denormalisation (IVDN) layer is then applied to produce the final compressed features $\{{\bm{\hat{P}}}_i\}_{i=2}^5$, which are fed to the task-specific tail network for downstream tasks, such as object detection and instance segmentation \cite{11165183}. The compressed machine-vision features also serve as the foundations for our human-vision compression, achieved by our fusion control network (FCN) as shall be introduced in the sequel.}

{To reconstruct high-fidelity images that align with human visual perception from low-bitrate machine-vision features, we leverage the diffusion prior from the pretrained stable diffusion model to restore missing visual details. To fully exploit the semantic and structural information embedded in machine vision features, we propose the novel FCN module. Unlike existing strategies by conditioning on diffusion models, FCN explicitly integrates multi-scale semantic features $\{{\bm{\hat{P}}}_2, {\bm{\hat{P}}}_3\}$ with extra colour cues ${\bm{z}}_{s}$ extracted from the encoder of the diffusion model. To further reduce the information volume of ${\bm{z}}_{s}$, we introduce an auxiliary compression network (ACN) to compress it. Those complementary cues are fused through a dedicated fusion module and combined with the noisy latent $\bm{z}_t$ to constitute a complete representation of both image semantics and details. The established representation is thus digested by a control module, which precisely and seamlessly guides the diffusion process for high-fidelity image reconstruction via our HVCN.}

\subsection{Variable-rate Feature Compression for Machine Vision}


{In our VFCN, we propose a simple yet effective normalisation strategy based on single input multiple output framework (SIMO) \cite{zhang2024hybrid} to enable variable-rate compression, which also enhances performance on downstream machine vision tasks. More specifically, in a fixed-rate compression model, the statistical distribution of the input features determines that of the transformed latent representation, which in turn governs the required quantisation granularity and the resulting bitrate. Therefore, by manipulating the distribution of the input features without modifying the quantisation scheme, it is able to realise variable-rate compression. Based on this, we propose a lightweight IVN layer that adjusts the statistical distribution of input features, enabling efficient variable-rate compression without requiring any modification to the quantisation mechanism. We may need to point out that existing studies on variable-rate compression for machine vision remain limited, and most approaches follow the paradigm of variable-rate image compression \cite{cui2021asymmetric, tong2023qvrf, duan2023qarv}, which typically introduce conditional encoders to adaptively weight the features after the encoder \cite{hossain2023flexible}. Aiming at rescaling the \textit{input} before the encoder, our variable-rate technique is simple yet effective, and different from existing methods that rescaling the \textit{output} after the encoder. Shifting from output to input levels preserves distributions for the entropy model and implicitly highlights task-oriented regions of interest during rescaling. }

{In other words, during the training phase, the IVN layer is expected to handle features on varying scales. To achieve this, we apply a local min-max normalisation at the batch level across multi-scale features $\{{\bm{P}}_i\}_{i=2}^5$:
\begin{equation}
\begin{aligned}
\bm{P}_0 &= \bm{P}_2\oplus\bm{P}_3\oplus\bm{P}_4\oplus\bm{P}_5 \\
\bar{\bm{P}}_{i} &= \frac{\bm{P}_i-\min(\bm{P}_0)}{\max(\bm{P}_0)-\min(\bm{P}_0)},
\end{aligned}
\end{equation}
where the minimum and maximum values are computed from the combined feature $\bm{P}_0$ of all scales within the batch and $\oplus$ denotes concatenation across batches. We use the normalized $\bar{\bm{P}}_2$ feature as the input to the SIMO network and obtain ${\bm{y}}_{p}$ and ${\bm{z}}_{p}$ by the transformation network (DR) and hyper encoder (HE) with parameter $\bm{\phi_g}$ and $\bm{\phi_h}$.These are then quantized to $\hat{\bm{y}}_{p}$ and $\hat{\bm{z}}_{p}$:
\begin{equation}
\label{eq:y_p}
\begin{aligned}
\hat{\bm{y}}_{p} &=Q({\bm{y}}_{p}) = Q(\mathrm{DR}(\bar{\bm{P}}_{2};\bm{\phi_g})) \\
\hat{\bm{z}}_{p} &=Q({\bm{z}}_{p}) = Q(\mathrm{HE}(\bm{y}_{p};\bm{\phi_h})),
\end{aligned}
\end{equation}
where $Q(\cdot)$ represents the quantisation. }

\begin{figure*}[htb]
   \begin{center}
   \includegraphics[width=1\linewidth]{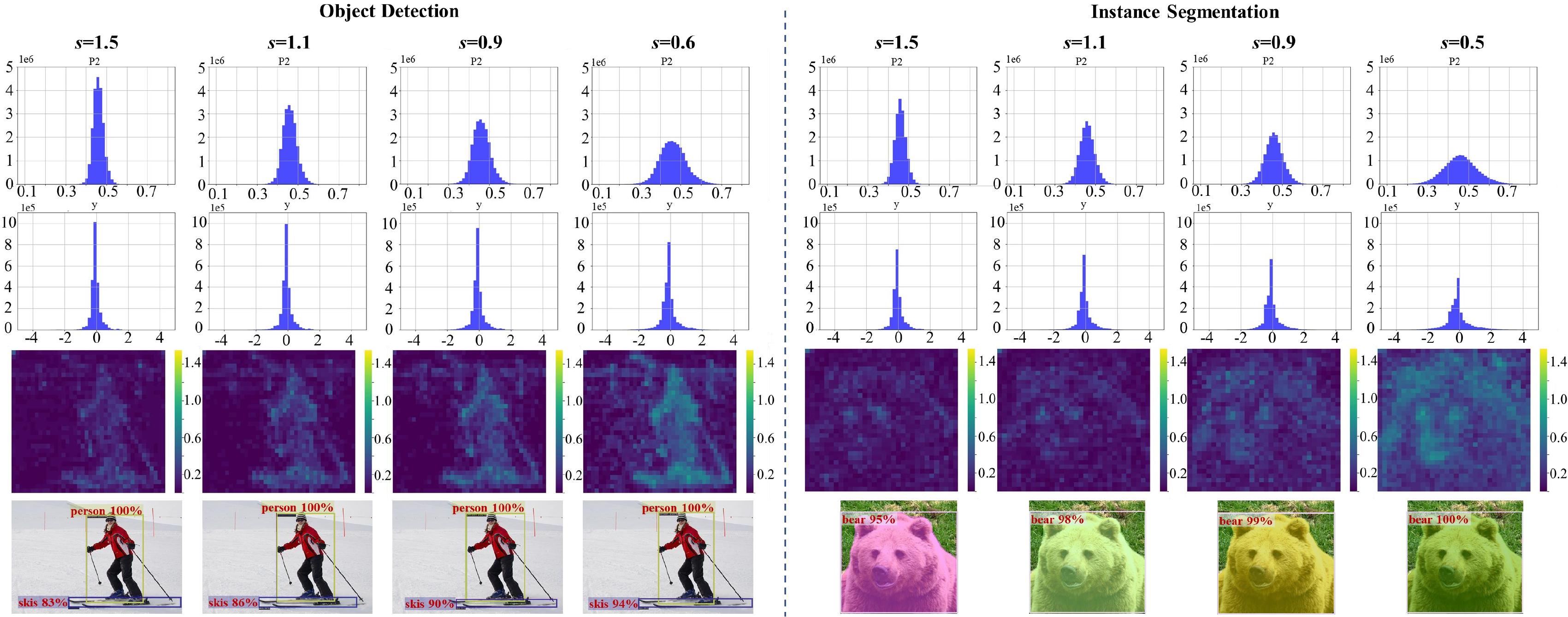}
   \end{center}
      \caption{{Visualization of feature distributions, bit allocation, object detection and instance segmentation results under four scaling factors. The first row illustrates the distribution of the scaled input $\bar{P}_2$ features, while the second row shows the distribution of the latent $\bm{y}_p$. The third row presents the bit allocation maps, computed by averaging the negative log-likelihood across channels. The final row displays the object detection and instance segmentation results corresponding to each scaling factor.}}
   \label{fig:var_analysis}
\end{figure*}  

{Then, four up-sample restoration (UR) modules with parameters $\{\bm{\theta_{s_i}}\}_{i=1}^4$ are employed to reconstruct the compressed latent representation $\bm{\hat{y}}_{p}$ to match the spatial dimensions of the original normalised features $\{\bar{\bm{P}}_i\}_{i=2}^5$, as follows:
\begin{equation}
\begin{aligned}
\tilde{\bm{P}}_{2} &= \mathrm{UR}_1(\bm{\hat{y}_{p}};\bm{\theta_{s_{1}}}),\quad \tilde{\bm{P}}_{3} = \mathrm{UR}_2(\bm{\hat{y}_{p}};\bm{\theta_{s_{2}}}), \\
\tilde{\bm{P}}_{4} &= \mathrm{UR}_3(\bm{\hat{y}_{p}};\bm{\theta_{s_{3}}}),\quad \tilde{\bm{P}}_{5} = \mathrm{UR}_4(\bm{\hat{y}_{p}};\bm{\theta_{s_{4}}}).
\end{aligned}  
\end{equation}
The prediction loss is defined by the mean squared error (MSE) to restrict the reconstruction quality for all the normalised features:
\begin{equation}
\label{eq:l_p}
\mathcal{L}_{{p}}=\sum_{i=2}^5\|{\tilde{\bm{P}}_i-\bar{\bm{P}}_i}\|_2^2.
\end{equation}}

{During the inference stage, we employ fixed global minimum and maximum values, denoted as $c_\mathrm{min}$ and $c_\mathrm{max}$, scaled by a factor $s$, as the IVN layer parameters:
\begin{equation}
\bar{\bm{P}}_{i} = \frac{\bm{P}_i-c_\mathrm{min}\cdot s}{c_\mathrm{max}\cdot s-c_\mathrm{min}\cdot s} = \frac{\nicefrac{\bm{P}_i}{s}-c_\mathrm{min}}{c_\mathrm{max}-c_\mathrm{min}}.
\end{equation}
This essentially differs from the training stage, where batch-dependent statistics are used. The decompressed $\{\tilde{\bm{P}}_i\}_{i=2}^5$ are denormalised to produce the final feature maps $\{\hat{\bm{P}}_i\}_{i=2}^5$ for subsequent vision tasks: 
\begin{equation}
\hat{\bm{P}}_{i} = [\tilde{\bm{P}}_{i}\cdot (c_\mathrm{max}-c_\mathrm{min}) + c_\mathrm{min}] \cdot s.
\end{equation}}

{Fig. \ref{fig:var_analysis} demonstrates that adjusting the input distribution of $\bar{\bm{P}}_2$ via the scaling factor $s$ effectively alters the distribution of $\bm{y}_{p}$, thereby enabling variable-rate compression. A broader distribution corresponds to more concentrated bit allocation maps (with more green regions), thus indicating higher bit rates and leading to improved detection accuracy. More importantly, unlike existing variable-rate  methods that explicitly manipulate the quantisation process to control bitrate, our approach achieves variable-rate compression by modulating the distribution of input features, providing a fundamentally different but effective strategy, which is also applicable to all types of feature compression network paradigm.}

\subsection{Fusing Diffusion Prior for Human Vision }

{After obtaining the compressed multi-scale features $\{\hat{\bm{P}}_i\}_{i=2}^5$ through our VFCN, we leverage a pre-trained stable diffusion model to reconstruct human-perceivable images. However, $\{\hat{\bm{P}}_i\}_{i=2}^5$ predominantly preserve structural and textural information from the original image, while lacking colour details. This limitation arises because tasks such as object detection and segmentation typically do not require accurate colour information. To address this issue, we introduce a lightweight auxiliary compression network (ACN) that complements the color bitstream to restore color fidelity. The colour features $\bm{z}_{{s}}$ are extracted from original image $\bm{x}$ using the encoder $\mathcal{E}(\cdot)$ from the stable diffusion model:
\begin{equation}
\label{eq:z_s}
\bm{z}_{{s}}=\mathcal{E}(\bm{x}), 
\end{equation}
which contains rich semantic information and aligns well with the generative latent space, thereby ensuring compatibility along with the image reconstruction process.}

{To further reduce the bitrate without increasing the overall network complexity, we develop a factorised entropy model to compress $\bm{z}_{{s}}$. More specifically, a latent encoder (LE) with parameter $\bm{\phi_{s}}$ is employed to transform $\bm{z}_{{s}}$ into a low-dimensional representation $\bm{y}_{{s}}$, which is subsequently reconstructed by a latent decoder (LD) with quantized $\hat{\bm{y}}_{s}$ and parameter $\bm{\theta_{s}}$, as follows,
\begin{equation}
\label{eq:y_s}
\begin{aligned}
\hat{\bm{y}}_{s} =Q&({\bm{y}}_{s})= Q(\mathrm{LE}({\bm{z}}_{{s}};\bm{\phi_{{s}}})), \\
\hat{\bm{z}}_{s} &= \mathrm{LD}(\hat{\bm{y}}_{s};\bm{\theta_{{s}}}).
\end{aligned}
\end{equation}
After obtaining all the auxiliary cues, we propose the fusion control network (FCN) that is composed of residual and attention blocks, to generate the fused features. To further enhance performance, we concatenate the fused features with the noisy latent variable $\bm{z}_t$, which has been verified to facilitate convergence by enabling our FCN network aware of the stochasticity at each diffusion timestep \cite{lin2024diffbir}. The final fusion is denoted as $\bm{c}_{f}$:
\begin{equation}
\label{eq:c_f}
\bm{c}_{f}=\mathrm{FCN}(\bm{\hat{P}}_2, \bm{\hat{P}}_3, \bm{\hat{z}}_{{s}})\oplus \bm{{z}}_{{t}}.
\end{equation}}

{Inspired by the ControlNet framework \cite{zhang2023adding}, we design our control module (CM) by creating a trainable copy of the stable diffusion (SD) encoding layers, which are connected through zero convolution layers $\mathcal{Z}(\cdot)$. Furthermore, to reduce inference latency and overall network complexity, we downscale the number of channels in the CM, significantly saving the training cost. We thus are able to obtain the estimated noise $\bm{\epsilon}_\theta$ at every timestep $\bm{t}$:
\begin{equation}
\label{eq:dif}
\bm{\epsilon}_\theta(\bm{z}_{t},\bm{t},\bm{c}_{f})=\mathrm{SD}(\mathcal{Z}(\mathrm{CM}(\bm{c}_{f})), \bm{{z}}_{t}).
\end{equation}
Following the standard spaced DDPM sampling strategy \cite{ho2020denoising}, the final denoised latent feature $\bm{z}_0$ is fed into the decoder $\mathcal{D}(\cdot)$ of the stable diffusion model to generate the reconstructed image $\hat{\bm{x}}$:
\begin{equation}
\label{eq:x_hat}
\bm{\hat{x}} = \mathcal{D}(\bm{z}_0).
\end{equation}}

\subsection{Rate-distortion Optimization}

{
\noindent\textbf{VFCN Training.} We follow the hyperprior compression framework, whereby the bitrate $\mathcal{L}_{r}$ is determined based on the low-dimensional feature $\bm{y}_{p}$ and the hyper feature $\bm{z}_{p}$ in (\ref{eq:y_p}) according to Shannon entropy theory:
\begin{equation}
\label{eq:l_r}
\mathcal{L}_{{r}} = \mathbb{E}[-\log_{2}p_{\bm{\hat{y}}_{p}}(Q(\bm{y}_{p}))] +\mathbb{E}[-\log_{2}p_{\bm{\hat{z}}_{p}}(Q(\bm{z}_{p}))],
\end{equation}
where recall that $Q(\cdot)$ denotes the quantisation operation. The VFCN is trained in an end-to-end manner using a loss function that combines the bit-rate loss and the feature reconstruction loss, as defined in (\ref{eq:l_p}). The hyperparameter $\lambda_{p}$ is used to balance the trade-off between compression rate and distortion. Notably, we adopt a single fixed value of $\lambda_{p}$ and achieve variable-rate compression solely by adjusting the scale factor, without retraining the model. The total loss $\mathcal{L}_\mathrm{mv}$ for the VFCN can be defined as: 
\begin{equation}
\label{eq:l_mv}
\mathcal{L}_{\mathrm{mv}}=\mathcal{L}_{{r}}+\lambda_{{p}}\mathcal{L}_{{p}}.
\end{equation}}

{
\noindent\textbf{HVCN Training.} It is necessary to optimise the noise predictor in latent diffusion model in advance. Given a set of conditions, including the time step $\bm{t}$ and the fusion information $\bm{c}_{f}$ from pre-trained VFCN, the diffusion model is trained to learn a noise estimator $\bm{\epsilon}_\theta$, typically implemented by a U-Net model \cite{ronneberger2015u}. The objective is to accurately predict the noise that has been added to the noisy latent representation $\bm{z}_{t}$:
\begin{equation}
\label{eq:l_eps}
\mathcal{L}_{{s}}=\mathbb{E}_{\bm{z}_{{s}},\bm{t},\bm{c}_{f},\bm{\epsilon}}\|\bm{\epsilon}-\bm{\epsilon}_\theta(\bm{z}_{t},\bm{t},\bm{c}_{f})\|_2^2.
\end{equation}
Since an additional bitstream containing colour information in (\ref{eq:y_s}) is introduced for human visual reconstruction, it is also necessary to impose a rate constraint on this stream, which is achieved by the following:
\begin{equation}
\label{eq:l_rc}
\mathcal{L}_{\mathrm{rs}} = \mathbb{E}[-\log_{2}p_{\bm{\hat{y}}_{s}}(Q(\bm{y}_{s}))].
\end{equation}
Moreover, we add a space alignment loss \cite{li2024towards} to force the content variables to align with the diffusion space, providing necessary constraints for optimization:
\begin{equation}
\label{eq:l_a}
\mathcal{L}_{{a}}=\|\bm{c}_{f}-\bm{z}_{{s}}\|_2^2.
\end{equation}
Therefore, the total loss for the HVCN is then formulated with hyperparameter $\lambda_{{a}}$ and $\lambda_{\mathrm{rs}}$ to control the trade-off between bit-rates and distortion:
\begin{equation}
\label{eq:l_hv}
\mathcal{L}_{\mathrm{hv}}=\mathcal{L}_{{s}} + \lambda_{{a}}\mathcal{L}_{{a}} + \lambda_{\mathrm{rs}}\mathcal{L}_{\mathrm{rs}}.
\end{equation}}

\section{Experimental Evaluations}

\subsection{Experimental Settings}

{
\noindent\textbf{Datasets.} For machine vision tasks, we trained the VFCN network of our method for the object detection and instance segmentation, based on COCO2017 dataset \cite{lin2014microsoft}. Upon this, we further employed the Flicker 2W dataset \cite{liu2020unified} to train HVCN networks for human vision. 
Each image was randomly cropped into 256$\times$256 for training machine vision tasks and 512$\times$512 for training human vision tasks.
We followed the setting of Adapt-ICMH \cite{li2024image} to evaluate the performance of two machine vision tasks on COCO2017-val dataset and use Kodak dataset \cite{kodakdataset} to evaluate the fidelity for human vision.}


\noindent\textbf{Baseline Methods.} 
To comprehensively evaluate the superiority of our Diff-FCMH method for human-machine collaborative compression, we compared with the existing state-of-the-art collaborative compression methods including Adapt-ICMH \cite{li2024image}, TransTIC \cite{chen2023transtic}, ICMH \cite{liu2023icmh} and CS model \cite{liu2022improving}. 
For the methods without their official implementations, we refer to the reproduced results reported in \cite{li2024image} for fair comparisons. 
To evaluate the efficiency of our model in compressing machine vision features, we further compared against the L-MSFC model \cite{kim2023end} and AFC \cite{zhang2024afc}, which are pure machine-vision-oriented methods for FCM. Furthermore, to demonstrate the superiority of our approach for human vision, we evaluated our method against two representative image compression approaches: VVC \cite{bross2021overview} and ELIC \cite{he2022elic}. To validate the efficiency of our IVN strategy, we compared with QVRF \cite{tong2023qvrf} and AG \cite{cui2021asymmetric} methods to demonstrate the advantages introduced by our method.

\noindent\textbf{Evaluation Metrics.} We employed bits per pixel (bpp) to evaluate the bit-rate cost during the compression. For machine vision tasks, we use mean average precision (mAP) with an Intersection of Union (IoU) threshold of 0.5, the standard metric to measure the performance for detection and segmentation tasks. To evaluate human vision fidelity, we measured the perceptual distortion by the widely applied metrics, including the learned perceptual image patch similarity (LPIPS) \cite{zhang2018unreasonable} and  natural image quality evaluator (NIQE) \cite{mittal2012making}. To compare the compression performance of different algorithms, we use the BD-BR (Bjøntegaard-Delta rate) \cite{bjontegaard2001calculation} as a measurement metric and calculate the BD-Metric.

 \begin{figure}[b]
   \begin{center}
   \includegraphics[width=1\linewidth]{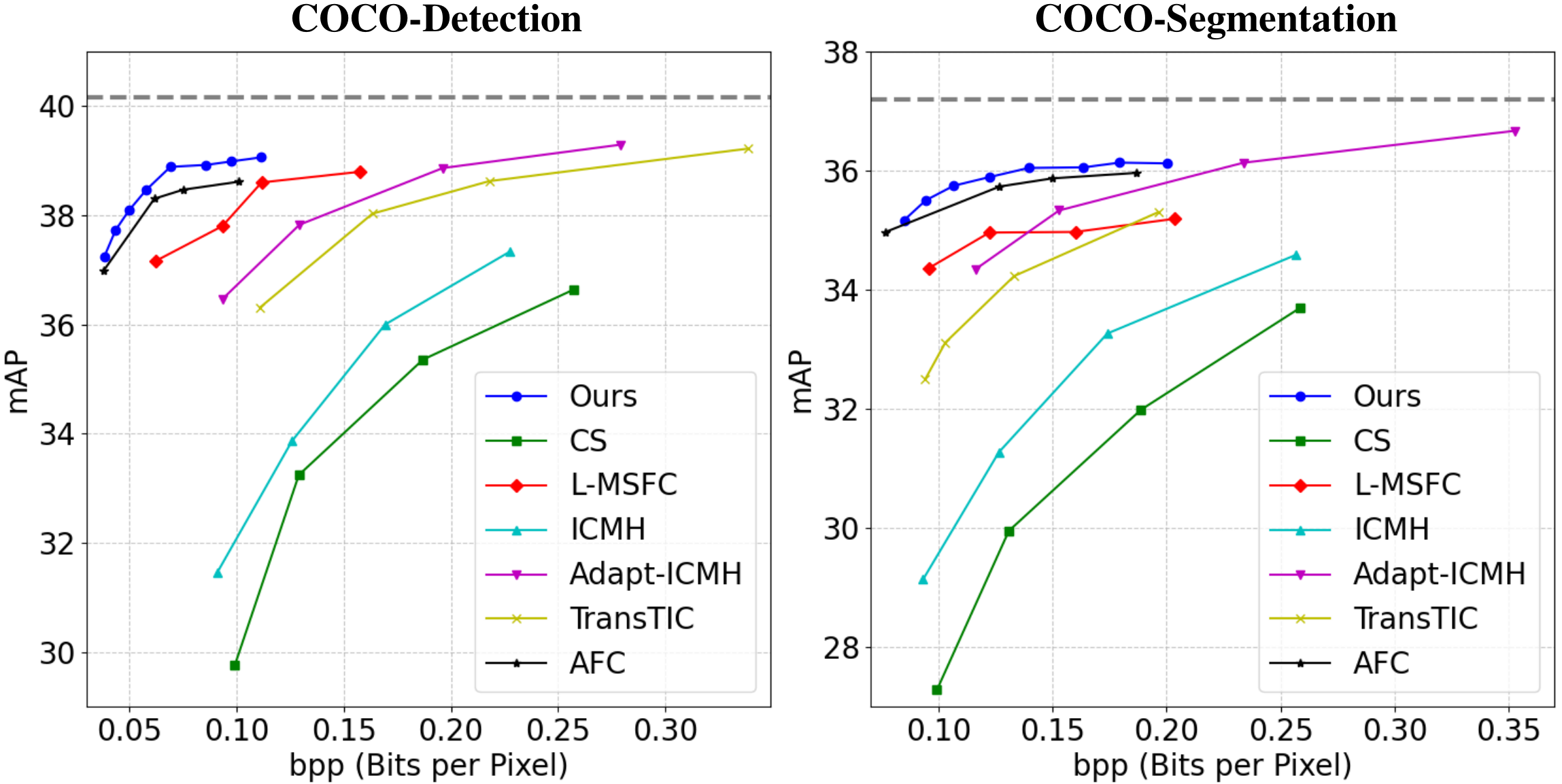}
   \end{center}
      \caption{Rate-mAP curves of our method and five comparing baseline methods for detection and segmentation tasks.}
   \label{fig:curve_mv}
\end{figure}

\noindent\textbf{Implementation Details.} 
We followed the same experimental configuration as Adapt-ICMH \cite{li2024image} using FPN-50 \cite{lin2017feature}, and employed the Faster R-CNN \cite{ren2015faster} and Mask R-CNN \cite{he2017mask} frameworks provided by \textit{Facebook Detectron2}, to extract multi-scale features $\{{\bm{P}}_i\}_{i=2}^5$ for object detection and segmentation tasks. For machine vision training, we set $\lambda_{{p}}=0.013$ for detection and $\lambda_{{p}}=0.025$ for segmentation. We utilized scale factor $s \in [0.4, 1.2]$ to achieve variable bitrates. For machine vision, compressed features are extracted from the detection task with fixed hyperparameters ($\lambda_{a}=2$, $\lambda_{\mathrm{rs}}=\{0.1, 0.5, 1, 3\}$) to support bitrate adaptation. Although our framework supports variable-rate compression for machine vision, reconstructing images for human vision still requires training multiple models at different rate levels.

\setlength{\tabcolsep}{2.6pt}
\begin{table*}[htbp]
\renewcommand{\arraystretch}{1.1}
\centering
\caption{Comparison of BD-BR, BD-mAP, BD-LPIPS (BD-L) and BD-NIQE (BD-N) for both machine vision and human vision. $\downarrow$ indicates that a lower score corresponds to better performance, $\uparrow$ indicates that higher is better, and bold font highlights the best values.}
{\small
\begin{tabular}{cc|cc|cc|ccc}
\toprule
\rowcolor[HTML]{EFEFEF} 
\multicolumn{2}{c|}{\textbf{Tasks}}  &\multicolumn{2}{c|}{{\textbf{COCO-Detection}}} &\multicolumn{2}{c|}{\textbf{COCO-Segmentation}} & \multicolumn{3}{c}{\textbf{Kodak-Compression}}\\ \hline
\multicolumn{2}{c|}{{Methods / Metrics}}& {{BD-BR (\%) $\downarrow$}} & {{BD-mAP $\uparrow$}} & {{BD-BR (\%) $\downarrow$}} & {{BD-mAP $\uparrow$}} & {BD-BR (\%) $\downarrow$} & {BD-L $\downarrow$} & {BD-N $\downarrow$}\\\hline
\rowcolor[HTML]{EFEFEF} 
\multicolumn{2}{c|}{VVC}  &0.00 & 0.00 & 0.00 & 0.00 & 0.00 & 0.00 & 0.00 \\
\multicolumn{2}{c|}{ELIC (CVPR, 2022)}  &-16.53 & 1.82 & -22.76 & 2.19 & 0.19 & 0.00 & 0.22 \\
\rowcolor[HTML]{EFEFEF} 
\multicolumn{2}{c|}{L-MSFC (TCSVT, 2023)}  &-92.93 & 16.29 & -91.91 & 12.92 & - & - & -\\
\multicolumn{2}{c|}{ICMH (ACMMM, 2023)}    & -70.07  & 9.95 & -71.95 & 9.37 & 30.93 &0.04 &0.32       \\
\rowcolor[HTML]{EFEFEF} 
\multicolumn{2}{c|}{TransTIC (ICCV, 2023)}  & -86.17  &10.46 &-85.90 & 12.44 &-14.02 &-0.01 &2.36 \\
\multicolumn{2}{c|}{AFC (ICME, 2024)}  &-95.42 & 18.57 & -95.01 & 14.71 & - & - & -\\
\rowcolor[HTML]{EFEFEF} 
\multicolumn{2}{c|}{Ada.-ICMH (ECCV, 2024)} &{-88.51} &{12.32} &{-91.15}& 10.33 &{-54.79} & -0.08 & 1.97  \\
\multicolumn{2}{c|}{Ours}   &{\textbf{-95.84}} &\textbf{18.89} &{\textbf{-95.22}} & {\textbf{14.89}} &{\textbf{-61.31}} &\textbf{-0.11} &\textbf{-1.41}\\
\bottomrule
\end{tabular}
}

\label{tab:table1}
\end{table*}

\begin{figure*}[htbp]
   \begin{center}
   \includegraphics[width=1\linewidth]{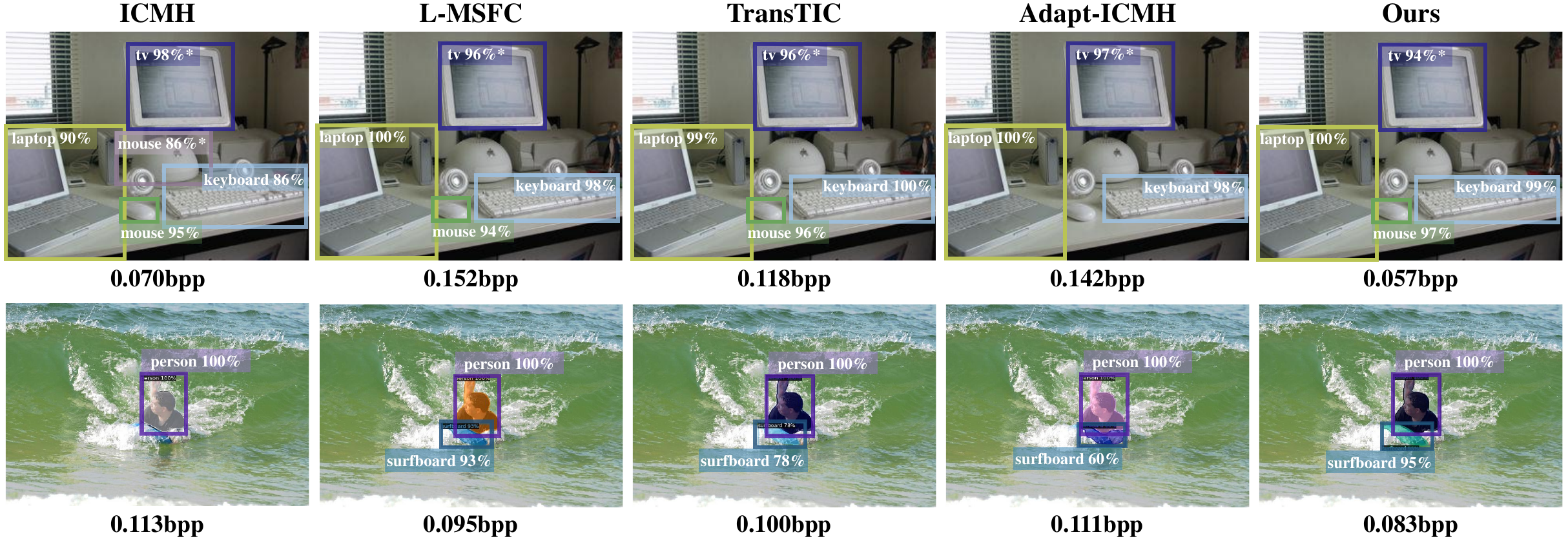}
   \end{center}
      \caption{Subjective results for machine vision tasks on the COCO dataset. Note that wrongly detected objects are annotated by the symbol $\star$. For correctly detected objects, a higher confidence score indicates a more reliable detection, whereas for incorrectly detected instances, a lower score reflects better suppression of false positives.}
   \label{fig:mv_sub}
\end{figure*}

\subsection{Evaluation for Machine Vision Compression}

We first compare the rate-accuracy performance of our methods with the state-of-the-art collaborative compression methods. Table \ref{tab:table1} reports the BD-rate and BD-mAP \cite{bjontegaard2001calculation} values averaged over detection and segmentation tasks for the COCO dataset, whereby the rate-accuracy (rate-mAP) curves are plotted in Fig. \ref{fig:curve_mv}. The compared fixed-rate compression methods are trained with four separate models, each corresponding to a different bitrate. In contrast, our variable-rate compression model achieves bitrate control through a single model by adjusting the scaling factor $s$. As shown in Fig. \ref{fig:curve_mv}, we sample eight points by varying the scale and considerably outperform all the existing baseline methods at all bit-rates, which achieves more than 95\% BD-BR saving against VVC anchor for the object detection and instance segmentation. The above results verify the superior performances of our variable-rate feature compression methods for machine vision. 

\begin{figure}[t]
   \begin{center}
   \includegraphics[width=1\linewidth]{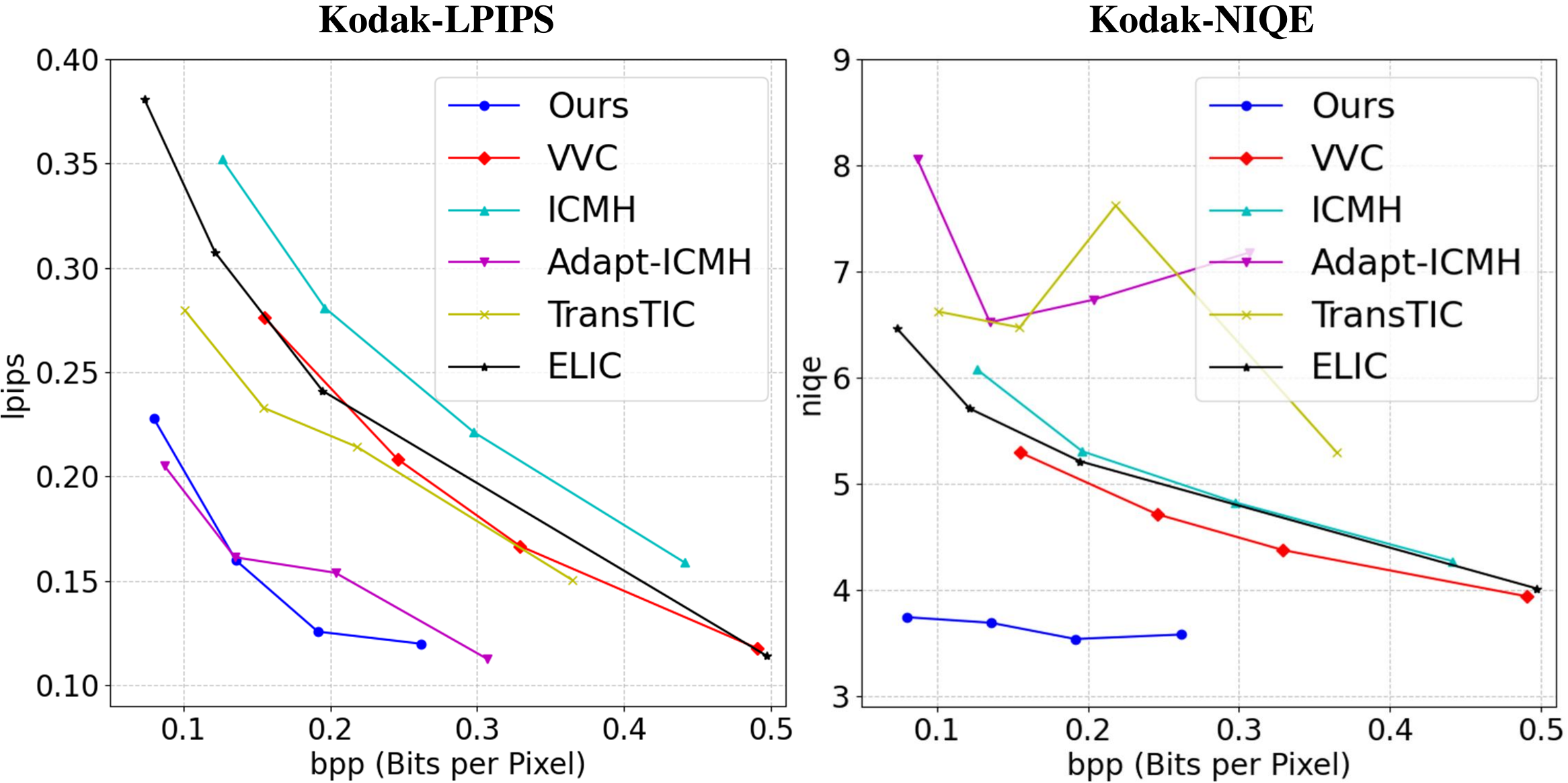}
   \end{center}
      \caption{Quantitative comparisons with the state-of-the-art methods in terms of perceptual quality (LPIPS and NIQE) on the Kodak dataset. Lower metrics indicate better quality.}
   \label{fig:curve_hm}
\end{figure}

We also illustrate qualitative results in Fig. \ref{fig:mv_sub}. For object detection, existing methods such as ICMH and Adapt-ICMH exhibit false positives and missed detections for objects (e.g., mouse). Although both L-MSFC and TransTIC correctly identify the object categories, their detection confidence scores are consistently lower than those produced by our Diff-FCHM. Moreover, since the COCO dataset does not include the category for desktop computers, all models incorrectly classify such instances as TV. Notably, our method assigns the lowest confidence scores to these misclassified cases, demonstrating the best reliability in abnormal cases. For the segmentation task, our method achieves the best confidence scores for small objects such as surfboard, demonstrating superior performance in fine-grained object recognition. Overall, across both detection and segmentation tasks, our approach achieves the best machine vision performance by the lowest bitrate.

\begin{figure*}[htbp]
   \begin{center}
   \includegraphics[width=1\linewidth]{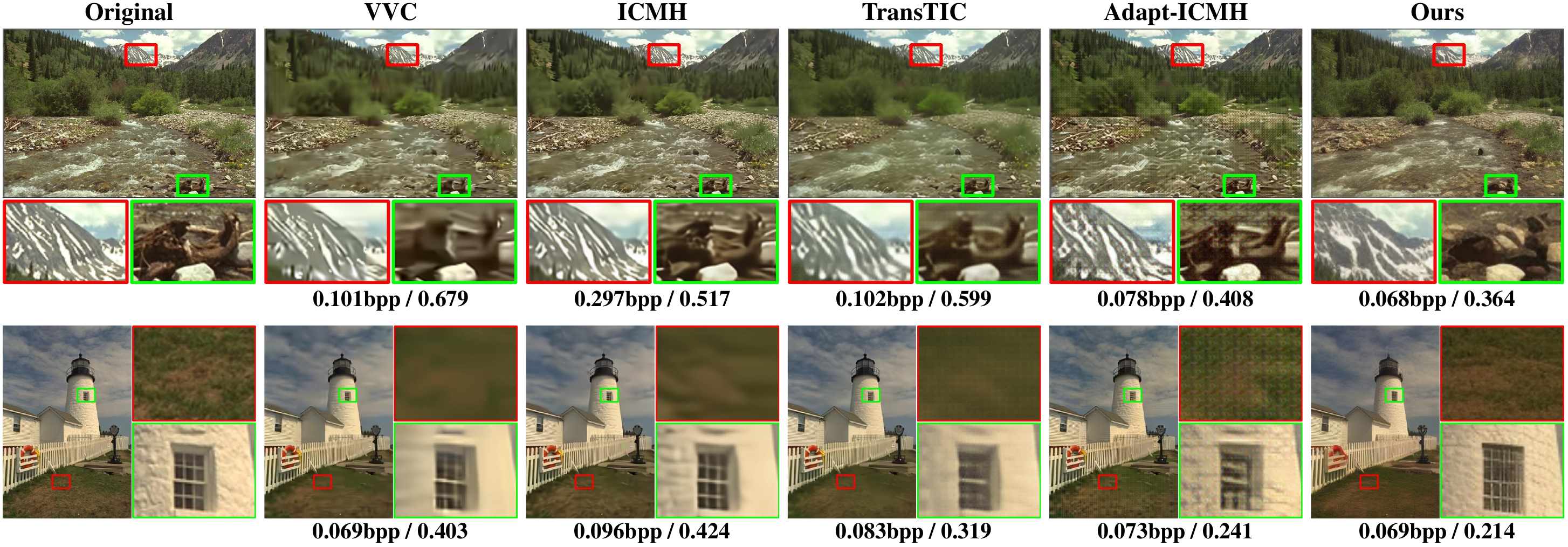}
   \end{center}
      \caption{Subjective results for human vision on the Kodak dataset under similar bit-rates. We also report the bpp and LPIPS metrics to quantify compression efficiency.}
   \label{fig:hv_sub}
\end{figure*}


\subsection{Evaluation for Human Vision Compression}

We illustrate the quantitative comparisons of our method with state-of-the-art collaborative approaches in Fig.\ref{fig:curve_hm}, in terms of perceptual quality measured by LPIPS and NIQE on the Kodak dataset. As shown in the left subfigure, our method consistently achieves lower LPIPS values across various bitrates, indicating better perceptual similarity with the uncompressed images. In the right subfigure, our method also obtains the lowest NIQE scores among all methods, suggesting superior visual quality from a non-reference perspective. It is observed that the NIQE scores of Adapt-ICMH and TransTIC are non-monotonic. This is expected, as their fine-tuning is guided by machine vision task losses rather than human perceptual quality, leading to compromise in visual fidelity from a human perspective. These results demonstrate the effectiveness of our method in preserving both task accuracy and perceptual quality under different compression levels. To further quantify the perceptual quality improvements, Table \ref{tab:table1} reports the BD-BR, BD-LPIPS, and BD-NIQE as distortion metrics. Our method achieves more than 61\% BD-BR savings (calculated by LPIPS) compared to the VVC anchor, demonstrating significant compression efficiency. Furthermore, our approach yields the lowest BD-LPIPS and BD-NIQE values, surpassing all competing methods in perceptual quality. These results confirm that our model achieves both better visual quality and more efficient rate-perception trade-off across bitrates. 

We also illustrate visual comparisons on two randomly selected Kodak images under similar bitrates, with LPIPS employed to measure perceptual fidelity, as shown in  Fig. \ref{fig:hv_sub}. Compared with the original image, VVC introduces strong blurring and oversmoothing artifacts in both texture and edge regions, leading to high LPIPS values. ICMH preserves more structures but suffers from significant detail loss and cartoon-like textures, especially in complex regions. Moreover, TransTIC achieves sharper edges than ICMH, but introduces heavy block artifacts and unnatural texture patterns, reducing visual realism. Adapt-ICMH further improves texture restoration but still struggles with unnatural noise and inconsistent structures, particularly in repetitive patterns. In contrast, our method achieves the most visually pleasing reconstructions with sharper details, faithful textures, and minimal artifacts, even at the lowest bitrates. For example, the grass textures in our results are closely similar to the ground-truth image. The superior performance demonstrates the ability of our model to generate perceptually accurate and visually coherent images with enhanced rate-distortion trade-offs.

\subsection{Evaluation for Variable-rate Mechanism}

\begin{figure}[htbp]
   \begin{center}
   \includegraphics[width=1\linewidth]{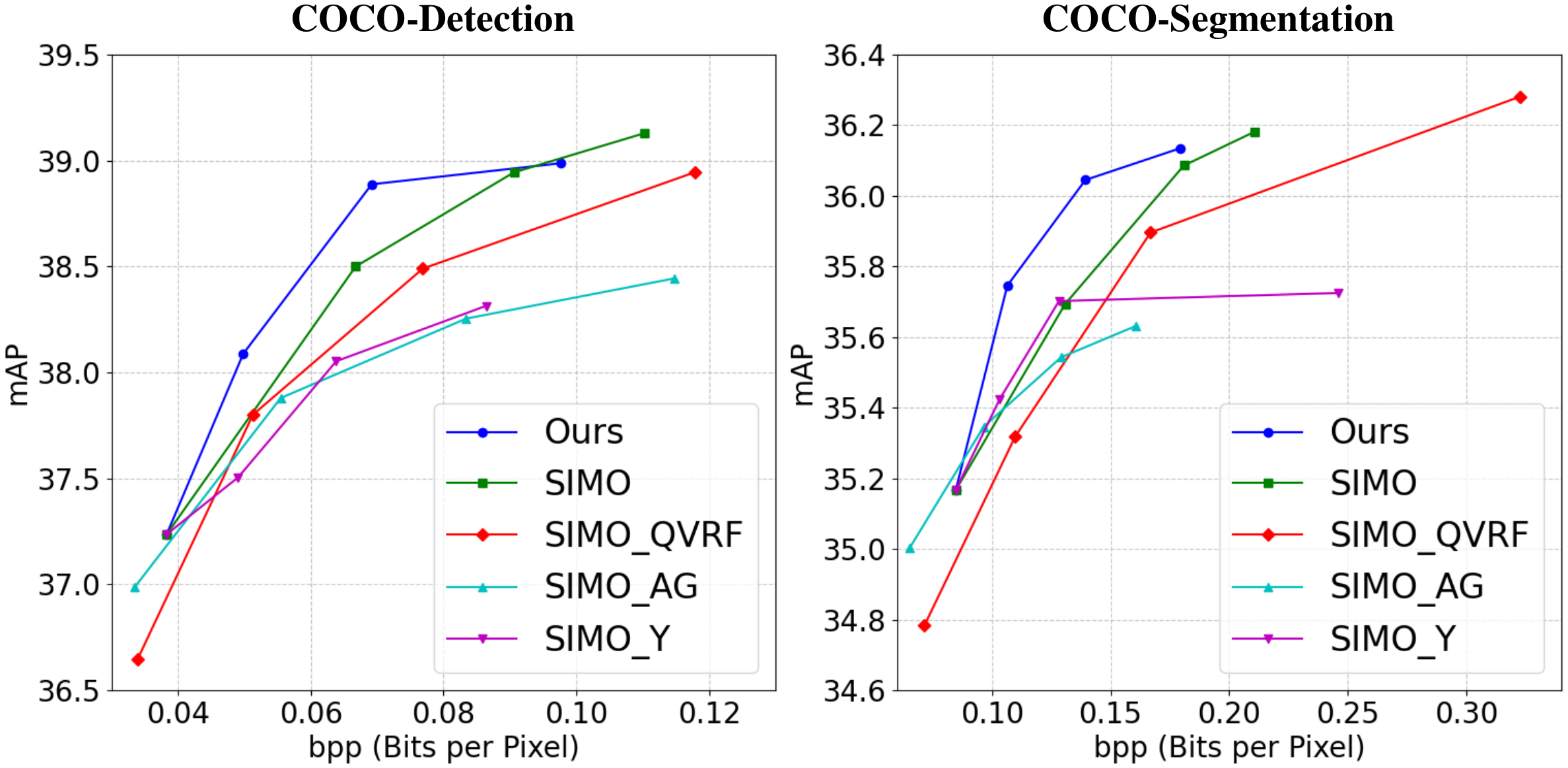}
   \end{center}
   \caption{Results of variable-rate methods with two baselines on detection and segmentation tasks.}
   \label{fig:curve_var}
\end{figure}

We compare our method with four representative baselines, pure SIMO \cite{zhang2024hybrid}, QVRF \cite{tong2023qvrf} based on SIMO, AG \cite{cui2021asymmetric} based on SIMO and an ablation variant with normalized output level (denoted by \text{SIMO-Y}). The original SIMO approach essentially trains multiple fixed-rate networks independently, whereas the other three methods train a single variable-rate network due to the feasibility of variable bit-rates. As shown in Fig. \ref{fig:curve_var} and Table \ref{tab:var-table}, our method consistently outperforms other methods across all bitrate levels for two tasks. 

\setlength{\tabcolsep}{3pt}
\begin{table}[b]
\renewcommand{\arraystretch}{1.2}
\centering
\caption{Quantitative results of variable bit-rate methods with four baselines on detection and segmentation tasks.}
{\small\begin{tabular}{c|cc|cc}
\toprule
\rowcolor[HTML]{EFEFEF}
\textbf{Tasks}  & \multicolumn{2}{c|}{\textbf{Detection}} &\multicolumn{2}{c}{\textbf{Segmentation}} \\ \hline
{Method/Metric} & {BD-BR$\downarrow$}   & {BD-mAP$\uparrow$}   & {BD-BR$\downarrow$}     & {BD-mAP$\uparrow$}    \\\hline
\rowcolor[HTML]{EFEFEF}SIMO                   & 0.00\%         & 0.00   & 0.00\% & 0.00\\
SIMO-QVRF              & 13.22\%      & -0.21   & 12.33\% & -0.15  \\
\rowcolor[HTML]{EFEFEF}SIMO-AG                & 17.55\%      & -0.38   & 2.79\% & -0.09\\
SIMO-Y                 & 19.68\%      & -0.36   & -0.11\% & -0.14\\
\rowcolor[HTML]{EFEFEF}Ours                   & \textbf{-9.49\%}       & \textbf{0.20}   & \textbf{-16.47\%}  & \textbf{0.22}\\
\bottomrule
\end{tabular}}
\label{tab:var-table}
\end{table}

More specifically, unlike existing methods that train with multiple $\lambda$ values and random learnable scales, our method benefits from a single $\lambda$ without additional learnable parameters, enabling faster convergence and improved results. For machine vision feature compression, the deficiency of in \text{SIMO-Y} from Fig. \ref{fig:curve_var}  demonstrates that designing at the input level is more effective than directly operating at the output level of the quantisation process. Moreover, compared with pure SIMO, our method surpasses the performance of the directly trained SIMO model. We attribute this advantage to the implicit normalisation strategy applied before compression, which acts as a soft filter that preserves important structural features (e.g., edges and contours) crucial for downstream visual tasks, while suppressing less informative, smooth background regions. This is also in accordance with Fig. \ref{fig:var_analysis}, in which the bitrate allocation maps for features normalized with different scale factors tend to allocate more bits to the detected target regions. This novel strategy maintains task-relevant information during compression, enabling better task performance even with reduced bitrate budgets. 

\section{Analysis and Discussions}

\subsection{Ablation Study for Mixed Condition}

\begin{figure}[t]
   \begin{center}
   \includegraphics[width=1\linewidth]{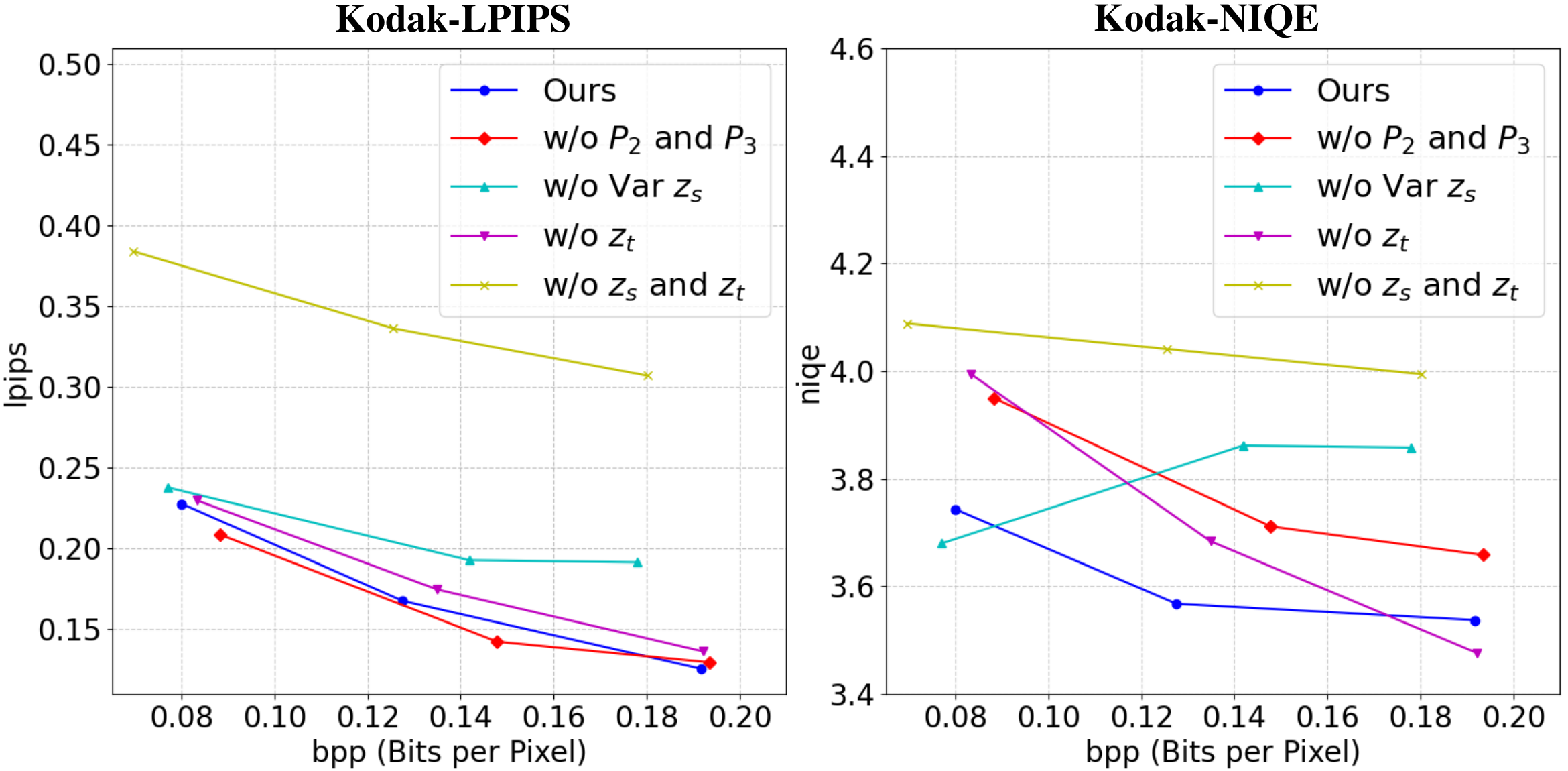}
   \end{center}
    \caption{Ablation studies on our Diff-FCMH for image reconstruction with different conditional latent.}
   \label{fig:curve_ab}
\end{figure}

To verify the effectiveness of the core components of our Diff-FCMH, we ablate different conditional latent $\bm{c}_{f}$ in (\ref{eq:c_f}), regarding the proposed FCN under various combinations of latent features as shown in Fig. \ref{fig:curve_ab}. We first verify that reconstructing images solely from machine vision features (denoted by \textit{w./o.} $\bm{z}_{{s}}$ and $\bm{z}_t$) leads to a significant drop in quality, especially under the LPIPS metric. Then, introducing the colour-sensitive latent $\bm{z}_{{s}}$ (denoted by \textit{w./o.} $\bm{z}_t$) notably improves the perceptual metrics, highlighting the importance of colour-related information for visual similarity. In contrast, excluding the temporal latent $\bm{z}_{t}$ causes a slight decline in both LPIPS and NIQE, indicating that temporal cues facilitate more accurate noise prediction during denoising. Additionally, replacing large-scale spatial features $\{\bm{\hat{P}}_2, \bm{\hat{P}}_3\}$ with smaller-scale features $\{\bm{\hat{P}}_4, \bm{\hat{P}}_5\}$ (denoted by \textit{w./o.} $\bm{\hat{P}}_2$ and $\bm{\hat{P}}_3$) results in comparable LPIPS but a clear increase in NIQE. This indicates that although local fidelity is preserved, global structure is impaired, emphasizing the necessity of large-scale features. Furthermore, we compare bitrate allocation strategies and find that using a fixed machine vision bitrate with variable $\bm{z}_{{s}}$ outperforms the reverse setting (denoted by \textit{w./o.} Var $\bm{z}_{{s}}$), underlining the importance of allocating more bits to colour-sensitive information for improved perceptual quality.

\begin{figure}[htbp]
   \begin{center}
   \includegraphics[width=1\linewidth]{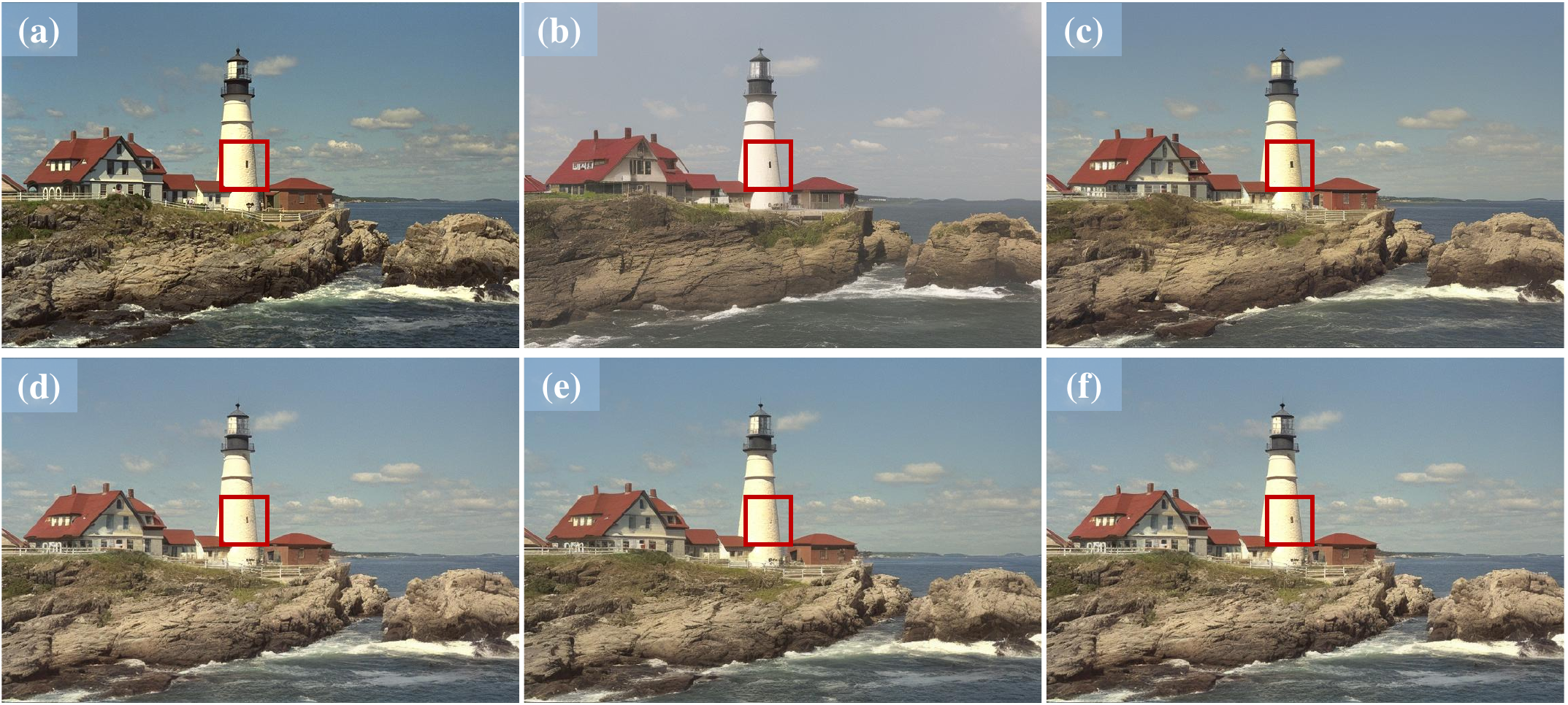}
   \end{center}
      \caption{Subjective results for human vision under similar bitrates: (a) Original image. (b) \textit{w./o.} $\bm{z}_{{s}}$ and $\bm{z}_t$. (c) \textit{w./o.} Var $\bm{z}_{{s}}$. (d) \textit{w./o.} $\bm{z}_t$. (e) \textit{w./o.} $\bm{\hat{P}}_2$ and $\bm{\hat{P}}_3$. (f) Proposed method.}
   \label{fig:sub_ab}
\end{figure}

 We further provide visual comparisons under similar bitrates in Fig. \ref{fig:sub_ab} to supplement the ablation study. As shown in Fig. \ref{fig:sub_ab}-(b), relying solely on machine vision features leads to noticeable colour distortions. In Figs. \ref{fig:sub_ab}-(c) and (d), ignoring $\bm{z}_t$ and $\bm{z}_{s}$ introduces slight noise in textured regions, indicating their contribution to fine-grained denoising. In Fig. \ref{fig:sub_ab}-(e), replacing large-scale features $\{\bm{\hat{P}}_2, \bm{\hat{P}}_3\}$ with small-scale features fails to recover structural details in the lighthouse area (as highlighted by the red box), underscoring the importance of larger receptive fields for capturing global context. 

\subsection{Effect of Denoising Steps}

\begin{figure*}[htbp]
   \begin{center}
   \includegraphics[width=1\linewidth]{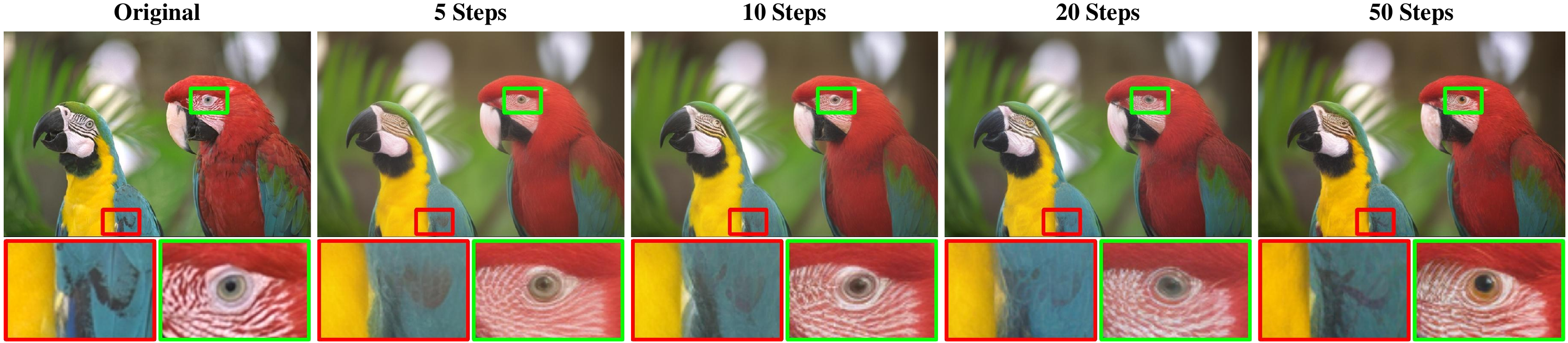}
   \end{center}
      \caption{Visual comparisons across 5, 10, 20, 50 denoising steps, which illustrate the progressive refinement of image quality under the same bit-rate. }
   \label{fig:sup_sample_sub}
\end{figure*}

Since the number of sampling steps in diffusion models directly affects both the generation speed and the perceptual quality of the output, we conduct further ablation studies to investigate its impact. As shown in Fig. \ref{fig:sup_sample_sub}, increasing the number of sampling steps generally leads to improved visual fidelity, including sharper textures and more accurate structural details. Within $50$ sampling steps, the generative model produces images with improved perceptual quality. However, due to generation uncertainty, it also introduces details that are not present in the original image, such as changes in eye colour. This reflects a common issue in generative model-based compression methods, where improved visual quality may compromise content accuracy.

\begin{figure}[htbp]
   \begin{center}
   \includegraphics[width=1\linewidth]{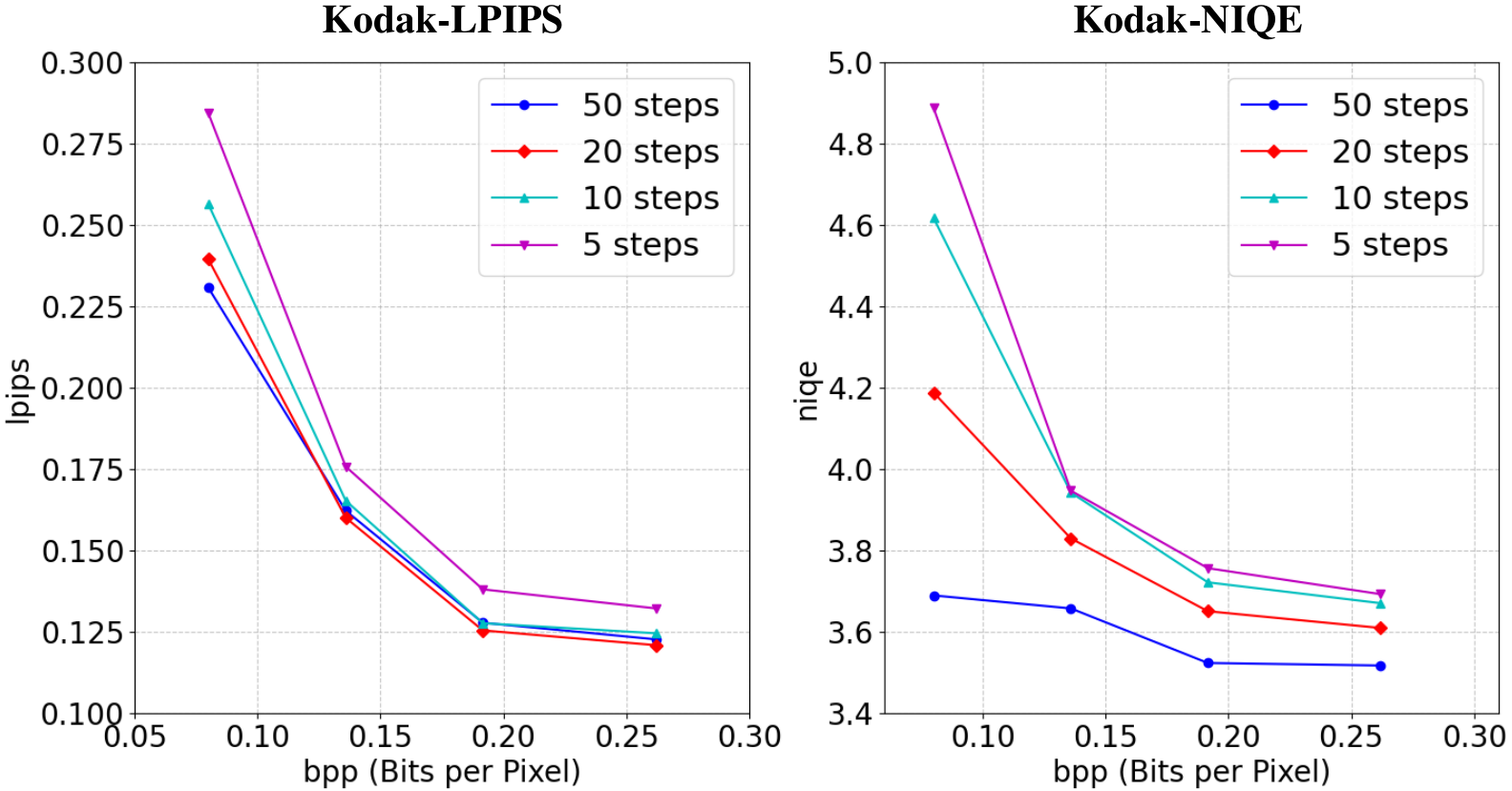}
   \end{center}
      \caption{Quantitative comparisons of denoising steps on the Kodak dataset for human vision.}
   \label{fig:sup_sample_curve}
\end{figure}

We conducted quantitative experiments to analyse the impact of sampling steps on perceptual quality under different bitrates. As shown in Fig. \ref{fig:sup_sample_curve}, we observe that LPIPS remains relatively stable when the sampling steps exceed $10$, while NIQE becomes stable after $20$ steps. However, achieving optimal quality requires $50$ sampling steps. The choice of sampling steps depends on hardware constraints and inference requirements.

\setlength{\tabcolsep}{8pt}
\begin{table}[ht]
\renewcommand{\arraystretch}{1.2}
\centering
\caption{Stability of LPIPS, NIQE, and MS-SSIM evaluated over compressed images generated with eight random seeds.}
\begin{tabular}{c|ccc}
\toprule
\rowcolor[HTML]{EFEFEF}\textbf{bpp} & \textbf{LPIPS} & \textbf{NIQE}  & \textbf{MS-SSIM} \\\hline
0.191        & 0.126 $\pm$ 8e-7 & 3.547 $\pm$ 9e-4 & 0.860 $\pm$ 5e-7   \\
\rowcolor[HTML]{EFEFEF}0.178        & 0.131 $\pm$ 1e-6 & 3.528 $\pm$ 5e-4 & 0.856 $\pm$ 6e-7  \\
\bottomrule
\end{tabular}
\label{tab:stable}
\end{table}

\subsection{Stability Analysis}

Due to the inherent randomness of diffusion models caused by varying random seeds, we conduct additional ablation studies to evaluate their stability. We calculate the deviation of images at 8 runs by different random seeds across low to high bitrates, and report the results in the Table \ref{tab:stable}. These results indicate that although diffusion models are inherently stochastic, the proposed method exhibits only negligible variations across different random seeds. The experimental findings demonstrate that our approach effectively suppresses quality fluctuations caused by the randomness of diffusion models, maintaining stable and consistent high-quality generation under varying conditions. Such stability is crucial for the practical deployment of compression methods, ensuring a reliable visual experience for users in all scenarios.

\section{Conclusion}

{In this paper, we have proposed a diffusion-prior assisted feature compression framework for both human and machine vision, named as Diff-FCHM. By compressing in the feature domain, the proposed variable-rate feature compression network (VFCN) exploits feature redundancy for efficient representation, together with an implicit variable normalisation (IVN) to handle feature sparsity and support variable bitrate encoding. For high-quality human visual reconstruction, we proposed the human vision compression network (HVCN), by introducing a fusion control network (FCN) module and auxiliary compression network (ACN) based on a pretrained stable diffusion backbone, enabling effective translation from compact visual features to perceptually accurate images. Extensive experiments have demonstrated that our method has significantly outperformed existing state-of-the-art approaches in human–machine collaborative compression. To the best of our knowledge, the proposed Diff-FCHM is the first framework that enables collaborative compression starting from machine vision features, which is an inherently challenging task due to the difficulty of reconstructing fine-grained visual content from high-level features. 
In future work, we plan to extend our approach to the video domain for broader applications in human–machine collaborative video compression.
}

\small{
\bibliographystyle{IEEEtran}

\bibliography{aaai2026}}

@article{kim2023end,
  title={End-to-end learnable multi-scale feature compression for VCM},
  author={Kim, Yeongwoong and Jeong, Hyewon and Yu, Janghyun and Kim, Younhee and Lee, Jooyoung and Jeong, Se Yoon and Kim, Hui Yong},
  journal={IEEE Transactions on Circuits and Systems for Video Technology},
  volume={34},
  number={5},
  pages={3156--3167},
  year={2023},
  publisher={IEEE}
}

@article{ren2015faster,
  title={Faster {R-CNN}: {T}owards real-time object detection with region proposal networks},
  author={Ren, S. and He, K. and Girshick, R. and Sun, J.},
  journal={NIPS},
  volume={28},
  year={2015}
}

@inproceedings{rippel2017real,
  title={Real-time adaptive image compression},
  author={Rippel, O. and Bourdev, L.},
  booktitle={ICML},
  pages={2922--2930},
  year={2017},
  organization={PMLR}
}

@article{sullivan2012overview,
  title={Overview of the high efficiency video coding (HEVC) standard},
  author={Sullivan, G. and Ohm, J. and Han, W. and Wiegand, T.},
  journal={IEEE Transactions on circuits and systems for video technology},
  volume={22},
  number={12},
  pages={1649--1668},
  year={2012},
  publisher={IEEE}
}

@article{bross2021overview,
  title={Overview of the versatile video coding ({VVC}) standard and its applications},
  author={Bross, B. and Wang, Y. and Ye, Y. and Liu, S. and Chen, J. and Sullivan, G. and Ohm, J.},
  journal={IEEE TCSVT},
  volume={31},
  number={10},
  pages={3736--3764},
  year={2021},
  publisher={IEEE}
}

@inproceedings{kang2022vcm,
	title={Feature Compression with resize in feature domain},
	author={Kang, J. and Jeong, H. and Bae, S. and Kim, H. and Kim, K. and Jeong, S.},
	booktitle={ISO/IEC JTC 1/SC 29/WG 2 m59537},
	pages={},
	year={2022}
}

@article{balle2018variational,
  title={Variational image compression with a scale hyperprior},
  author={Ball{\'e}, J. and Minnen, D. and Singh, S. and Hwang, S. and Johnston, N.},
  journal={arXiv preprint arXiv:1802.01436},
  year={2018}
}

@inproceedings{cheng2020learned,
  title={Learned image compression with discretized gaussian mixture likelihoods and attention modules},
  author={Cheng, Z. and Sun, H. and Takeuchi, M. and Katto, J.},
  booktitle={IEEE CVPR},
  pages={7939--7948},
  year={2020}
}

@inproceedings{he2017mask,
  title={Mask {R-CNN}},
  author={He, K. and Gkioxari, G. and Doll{\'a}r, P. and Girshick, R.},
  booktitle={IEEE ICCV},
  pages={2961--2969},
  year={2017}
}

@inproceedings{chen2023residual,
  title={Residual based hierarchical feature compression for multi-task machine vision},
  author={Chen, Chaoran and Xu, Mai and Li, Shengxi and Liu, Tie and Qiao, Minglang and Lv, Zhuoyi},
  booktitle={2023 IEEE International Conference on Multimedia and Expo (ICME)},
  pages={1463--1468},
  year={2023},
  organization={IEEE}
}

@inproceedings{CTCstandard,
	title={Proposed Common Test Conditions for Feature Compression for Video Coding for Machines},
	author={C. Rosewarne},
	booktitle={ISO/IEC JTC 1/SC 29/WG 4 m65749},
	pages={},
	year={2023}
}

@inproceedings{CTTCstandard,
	title={Proposed Common training conditions},
	author={R. Nguyen, C. Rosewarne},
	booktitle={ISO/IEC JTC 1/SC 29/WG 2 m65248},
	pages={},
	year={2023}
}

@inproceedings{lin2017feature,
  title={Feature pyramid networks for object detection},
  author={Lin, Tsung-Yi and Doll{\'a}r, Piotr and Girshick, Ross and He, Kaiming and Hariharan, Bharath and Belongie, Serge},
  booktitle={IEEE CVPR},
  pages={2117--2125},
  year={2017}
}

@article{li2024human,
  title={Human-Machine Collaborative Image and Video Compression: A Survey},
  author={Li, Huanyang and Zhang, Xinfeng and Wang, Shiqi and Wang, Shanshe and Pan, Jingshan and others},
  journal={APSIPA Transactions on Signal and Information Processing},
  volume={13},
  number={6},
  year={2024},
  publisher={Now Publishers, Inc.}
}

@inproceedings{lin2014microsoft,
  title={Microsoft coco: Common objects in context},
  author={Lin, Tsung-Yi and Maire, Michael and Belongie, Serge and Hays, James and Perona, Pietro and Ramanan, Deva and Doll{\'a}r, Piotr and Zitnick, C Lawrence},
  booktitle={Computer vision--ECCV 2014: 13th European conference, zurich, Switzerland, September 6-12, 2014, proceedings, part v 13},
  pages={740--755},
  year={2014},
  organization={Springer}
}

@article{liu2020unified,
  title={A unified end-to-end framework for efficient deep image compression},
  author={Liu, Jiaheng and Lu, Guo and Hu, Zhihao and Xu, Dong},
  journal={arXiv preprint arXiv:2002.03370},
  year={2020}
}

@inproceedings{li2024image,
  title={Image compression for machine and human vision with spatial-frequency adaptation},
  author={Li, Han and Li, Shaohui and Ding, Shuangrui and Dai, Wenrui and Cao, Maida and Li, Chenglin and Zou, Junni and Xiong, Hongkai},
  booktitle={European Conference on Computer Vision},
  pages={382--399},
  year={2024},
  organization={Springer}
}

@inproceedings{zhang2018unreasonable,
  title={The unreasonable effectiveness of deep features as a perceptual metric},
  author={Zhang, Richard and Isola, Phillip and Efros, Alexei A and Shechtman, Eli and Wang, Oliver},
  booktitle={Proceedings of the IEEE conference on computer vision and pattern recognition},
  pages={586--595},
  year={2018}
}

@article{li2024towards,
  title={Towards Extreme Image Compression with Latent Feature Guidance and Diffusion Prior},
  author={Li, Zhiyuan and Zhou, Yanhui and Wei, Hao and Ge, Chenyang and Jiang, Jingwen},
  journal={IEEE Transactions on Circuits and Systems for Video Technology},
  year={2024},
  publisher={IEEE}
}

@article{cao2023slimmable,
  title={Slimmable multi-task image compression for human and machine vision},
  author={Cao, Jiangzhong and Yao, Ximei and Zhang, Huan and Jin, Jian and Zhang, Yun and Ling, Bingo Wing-Kuen},
  journal={IEEE Access},
  volume={11},
  pages={29946--29958},
  year={2023},
  publisher={IEEE}
}

@inproceedings{liu2023icmh,
  title={Icmh-net: Neural image compression towards both machine vision and human vision},
  author={Liu, Lei and Hu, Zhihao and Chen, Zhenghao and Xu, Dong},
  booktitle={Proceedings of the 31st ACM International Conference on Multimedia},
  pages={8047--8056},
  year={2023}
}

@inproceedings{chen2023transtic,
  title={Transtic: Transferring transformer-based image compression from human perception to machine perception},
  author={Chen, Yi-Hsin and Weng, Ying-Chieh and Kao, Chia-Hao and Chien, Cheng and Chiu, Wei-Chen and Peng, Wen-Hsiao},
  booktitle={Proceedings of the IEEE/CVF International Conference on Computer Vision},
  pages={23297--23307},
  year={2023}
}

@inproceedings{liu2022improving,
  title={Improving multiple machine vision tasks in the compressed domain},
  author={Liu, Jinming and Sun, Heming and Katto, Jiro},
  booktitle={2022 26th International Conference on Pattern Recognition (ICPR)},
  pages={331--337},
  year={2022},
  organization={IEEE}
}

@article{choi2022scalable,
  title={Scalable image coding for humans and machines},
  author={Choi, Hyomin and Baji{\'c}, Ivan V},
  journal={IEEE Transactions on Image Processing},
  volume={31},
  pages={2739--2754},
  year={2022},
  publisher={IEEE}
}

@inproceedings{zhang2024afc,
  title={Afc: Asymmetrical Feature Coding for Multi-Task Machine Intelligence},
  author={Zhang, Yuan and Wang, Hanming and Li, Yunlong and Yu, Lu},
  booktitle={2024 IEEE International Conference on Multimedia and Expo Workshops (ICMEW)},
  pages={1--6},
  year={2024},
  organization={IEEE}
}

@inproceedings{zhang2024hybrid,
  title={Hybrid Single Input and Multiple Output Method For Compressing Features Towards Machine Vision Tasks},
  author={Zhang, Zifu and Li, Shengxi and Liu, Tie and Xu, Mai and Xu, Tao and Guan, Zhenyu and Lv, Zhuoyi},
  booktitle={2024 IEEE International Conference on Image Processing (ICIP)},
  pages={1870--1876},
  year={2024},
  organization={IEEE}
}

@inproceedings{guo2023toward,
  title={Toward scalable image feature compression: a content-adaptive and diffusion-based approach},
  author={Guo, Sha and Chen, Zhuo and Zhao, Yang and Zhang, Ning and Li, Xiaotong and Duan, Lingyu},
  booktitle={Proceedings of the 31st ACM International Conference on Multimedia},
  pages={1431--1442},
  year={2023}
}

@inproceedings{zhangall,
  title={All-in-One Image Coding for Joint Human-Machine Vision with Multi-Path Aggregation},
  author={Zhang, Xu and Guo, Peiyao and Lu, Ming and Ma, Zhan},
  booktitle={The Thirty-eighth Annual Conference on Neural Information Processing Systems},
  year={2024}
}

@article{wu2024scalable,
  title={Scalable image coding with enhancement features for human and machine},
  author={Wu, Ying and An, Ping and Yang, Chao and Huang, XinPeng},
  journal={Multimedia Systems},
  volume={30},
  number={2},
  pages={77},
  year={2024},
  publisher={Springer}
}

@article{yan2021sssic,
  title={SSSIC: semantics-to-signal scalable image coding with learned structural representations},
  author={Yan, Ning and Gao, Changsheng and Liu, Dong and Li, Houqiang and Li, Li and Wu, Feng},
  journal={IEEE Transactions on Image Processing},
  volume={30},
  pages={8939--8954},
  year={2021},
  publisher={IEEE}
}

@inproceedings{datta2022low,
  title={A low-complexity approach to rate-distortion optimized variable bit-rate compression for split dnn computing},
  author={Datta, Parual and Ahuja, Nilesh and Somayazulu, V Srinivasa and Tickoo, Omesh},
  booktitle={2022 26th International Conference on Pattern Recognition (ICPR)},
  pages={182--188},
  year={2022},
  organization={IEEE}
}

@inproceedings{hossain2023flexible,
  title={Flexible variable-rate image feature compression for edge-cloud systems},
  author={Hossain, Md Adnan Faisal and Duan, Zhihao and Huang, Yuning and Zhu, Fengqing},
  booktitle={2023 IEEE International Conference on Multimedia and Expo Workshops (ICMEW)},
  pages={182--187},
  year={2023},
  organization={IEEE}
}

@inproceedings{cui2021asymmetric,
  title={Asymmetric gained deep image compression with continuous rate adaptation},
  author={Cui, Ze and Wang, Jing and Gao, Shangyin and Guo, Tiansheng and Feng, Yihui and Bai, Bo},
  booktitle={Proceedings of the IEEE/CVF Conference on Computer Vision and Pattern Recognition},
  pages={10532--10541},
  year={2021}
}

@inproceedings{tong2023qvrf,
  title={Qvrf: A quantization-error-aware variable rate framework for learned image compression},
  author={Tong, Kedeng and Wu, Yaojun and Li, Yue and Zhang, Kai and Zhang, Li and Jin, Xin},
  booktitle={2023 IEEE International Conference on Image Processing (ICIP)},
  pages={1310--1314},
  year={2023},
  organization={IEEE}
}

@inproceedings{lu2019dvc,
  title={Dvc: An end-to-end deep video compression framework},
  author={Lu, Guo and Ouyang, Wanli and Xu, Dong and Zhang, Xiaoyun and Cai, Chunlei and Gao, Zhiyong},
  booktitle={Proceedings of the IEEE/CVF conference on computer vision and pattern recognition},
  pages={11006--11015},
  year={2019}
}

@article{goodfellow2020generative,
  title={Generative adversarial networks},
  author={Goodfellow, Ian and Pouget-Abadie, Jean and Mirza, Mehdi and Xu, Bing and Warde-Farley, David and Ozair, Sherjil and Courville, Aaron and Bengio, Yoshua},
  journal={Communications of the ACM},
  volume={63},
  number={11},
  pages={139--144},
  year={2020},
  publisher={ACM New York, NY, USA}
}

@article{ho2020denoising,
  title={Denoising diffusion probabilistic models},
  author={Ho, Jonathan and Jain, Ajay and Abbeel, Pieter},
  journal={Advances in neural information processing systems},
  volume={33},
  pages={6840--6851},
  year={2020}
}

@article{mittal2012making,
  title={Making a “completely blind” image quality analyzer},
  author={Mittal, Anish and Soundararajan, Rajiv and Bovik, Alan C},
  journal={IEEE Signal processing letters},
  volume={20},
  number={3},
  pages={209--212},
  year={2012},
  publisher={IEEE}
}

@inproceedings{lin2024diffbir,
  title={Diffbir: Toward blind image restoration with generative diffusion prior},
  author={Lin, Xinqi and He, Jingwen and Chen, Ziyan and Lyu, Zhaoyang and Dai, Bo and Yu, Fanghua and Qiao, Yu and Ouyang, Wanli and Dong, Chao},
  booktitle={European Conference on Computer Vision},
  pages={430--448},
  year={2024},
  organization={Springer}
}

@article{duan2023qarv,
  title={Qarv: Quantization-aware resnet vae for lossy image compression},
  author={Duan, Zhihao and Lu, Ming and Ma, Jack and Huang, Yuning and Ma, Zhan and Zhu, Fengqing},
  journal={IEEE Transactions on Pattern Analysis and Machine Intelligence},
  volume={46},
  number={1},
  pages={436--450},
  year={2023},
  publisher={IEEE}
}

@inproceedings{ronneberger2015u,
  title={U-net: Convolutional networks for biomedical image segmentation},
  author={Ronneberger, Olaf and Fischer, Philipp and Brox, Thomas},
  booktitle={Medical image computing and computer-assisted intervention--MICCAI 2015: 18th international conference, Munich, Germany, October 5-9, 2015, proceedings, part III 18},
  pages={234--241},
  year={2015},
  organization={Springer}
}

@inproceedings{zhang2023adding,
  title={Adding conditional control to text-to-image diffusion models},
  author={Zhang, Lvmin and Rao, Anyi and Agrawala, Maneesh},
  booktitle={Proceedings of the IEEE/CVF international conference on computer vision},
  pages={3836--3847},
  year={2023}
}

@article{bjontegaard2001calculation,
  title={Calculation of average PSNR differences between RD-curves},
  author={Bjontegaard, Gisle},
  journal={ITU SG16 Doc. VCEG-M33},
  year={2001}
}

@inproceedings{he2022elic,
  title={Elic: Efficient learned image compression with unevenly grouped space-channel contextual adaptive coding},
  author={He, Dailan and Yang, Ziming and Peng, Weikun and Ma, Rui and Qin, Hongwei and Wang, Yan},
  booktitle={Proceedings of the IEEE/CVF Conference on Computer Vision and Pattern Recognition},
  pages={5718--5727},
  year={2022}
}

@article{li2024humanJESTA,
  title={Human-machine collaborative image compression method based on implicit neural representations},
  author={Li, Huanyang and Zhang, Xinfeng},
  journal={IEEE Journal on Emerging and Selected Topics in Circuits and Systems},
  year={2024},
  publisher={IEEE}
}

@article{he2024learned,
  title={Learned Image Coding for Human-Machine Collaborative Optimization},
  author={He, Jingbo and He, Xiaohai and Xiong, Shuhua and Chen, Honggang},
  journal={IEEE Transactions on Broadcasting},
  year={2024},
  publisher={IEEE}
}

@article{duan2015overview,
  title={Overview of the MPEG-CDVS standard},
  author={Duan, Ling-Yu and Chandrasekhar, Vijay and Chen, Jie and Lin, Jie and Wang, Zhe and Huang, Tiejun and Girod, Bernd and Gao, Wen},
  journal={IEEE Transactions on Image Processing},
  volume={25},
  number={1},
  pages={179--194},
  year={2015},
  publisher={IEEE}
}

@article{mao2023scalable,
  title={Scalable face image coding via stylegan prior: Toward compression for human-machine collaborative vision},
  author={Mao, Qi and Wang, Chongyu and Wang, Meng and Wang, Shiqi and Chen, Ruijie and Jin, Libiao and Ma, Siwei},
  journal={IEEE Transactions on Image Processing},
  volume={33},
  pages={408--422},
  year={2023},
  publisher={IEEE}
}

@inproceedings{hu2020towards,
  title={Towards coding for human and machine vision: A scalable image coding approach},
  author={Hu, Yueyu and Yang, Shuai and Yang, Wenhan and Duan, Ling-Yu and Liu, Jiaying},
  booktitle={2020 IEEE International Conference on Multimedia and Expo (ICME)},
  pages={1--6},
  year={2020},
  organization={IEEE}
}

@inproceedings{bai2022towards,
  title={Towards end-to-end image compression and analysis with transformers},
  author={Bai, Yuanchao and Yang, Xu and Liu, Xianming and Jiang, Junjun and Wang, Yaowei and Ji, Xiangyang and Gao, Wen},
  booktitle={Proceedings of the AAAI conference on artificial intelligence},
  volume={36},
  pages={104--112},
  year={2022}
}

@inproceedings{kodakdataset,
  title={Kodak lossless true color image suite},
  author={Eastman Kodak Company},
  booktitle={http://r0k.us/graphics/kodak/},
  volume={},
  number={1},
  pages={},
  year={2013}
}

@inproceedings{careil2023towards,
  title={Towards image compression with perfect realism at ultra-low bitrates},
  author={Careil, Marlene and Muckley, Matthew J and Verbeek, Jakob and Lathuili{\`e}re, St{\'e}phane},
  booktitle={The Twelfth International Conference on Learning Representations},
  year={2024}
}

@article{akbari2021learned,
  title={Learned multi-resolution variable-rate image compression with octave-based residual blocks},
  author={Akbari, Mohammad and Liang, Jie and Han, Jingning and Tu, Chengjie},
  journal={IEEE Transactions on Multimedia},
  volume={23},
  pages={3013--3021},
  year={2021},
  publisher={IEEE}
}

@article{tu2025msinn,
  author    = {Tu, Hanyue and Wu, Siqi and Li, Li and Zhou, Wengang and Li, Houqiang},
  title     = {Multi-Scale Invertible Neural Network for Wide-Range Variable-Rate Learned Image Compression},
  journal   = {IEEE Transactions on Multimedia},
  year      = {2025},
}

@inproceedings{choi2019variable,
  title={Variable rate deep image compression with a conditional autoencoder},
  author={Choi, Yoojin and El-Khamy, Mostafa and Lee, Jungwon},
  booktitle={Proceedings of the IEEE/CVF international conference on computer vision},
  pages={3146--3154},
  year={2019}
}

@article{zhang2025HSC,
  author={Li, Shengxi and Zhang, Zifu and Xu, Mai and Jiang, Lai and Liu, Yufan and Zhu, Ce},
  journal={IEEE Transactions on Image Processing}, 
  title={Hierarchical Semantic Compression for Consistent Image Semantic Restoration}, 
  year={2025},
  volume={34},
  number={},
  pages={6767-6782},
  doi={10.1109/TIP.2025.3618379}}

@article{song2020denoising,
  title={Denoising diffusion implicit models},
  author={Song, Jiaming and Meng, Chenlin and Ermon, Stefano},
  journal={arXiv preprint arXiv:2010.02502},
  year={2020}
}

@inproceedings{agustsson2023multi,
  title={Multi-realism image compression with a conditional generator},
  author={Agustsson, Eirikur and Minnen, David and Toderici, George and Mentzer, Fabian},
  booktitle={IEEE/CVF CVPR},
  pages={22324--22333},
  year={2023}
}

@article{mentzer2020high,
  title={High-fidelity generative image compression},
  author={Mentzer, Fabian and Toderici, George D and Tschannen, Michael and Agustsson, Eirikur},
  journal={Advances in NIPS},
  volume={33},
  pages={11913--11924},
  year={2020}
}

@inproceedings{lei2023text+sketch,
  title={Text+ Sketch: Image Compression at Ultra Low Rates},
  author={Lei, Eric and Uslu, Yi\u{g}it Berkay and Hassani, Hamed and Bidokhti, Shirin Saeedi},
  booktitle={ICML 2023 Workshop on Neural Compression: From Information Theory to Applications},
  year={2023}
}

@article{li2024misc,
  title={Misc: Ultra-low bitrate image semantic compression driven by large multimodal model},
  author={Li, Chunyi and Lu, Guo and Feng, Donghui and Wu, Haoning and Zhang, Zicheng and Liu, Xiaohong and Zhai, Guangtao and Lin, Weisi and Zhang, Wenjun},
  journal={IEEE Transactions on Image Processing},
  year={2024},
  publisher={IEEE}
}

@article{zhang2025continuous,
  title={Continuous Patch Stitching for Block-wise Image Compression},
  author={Zhang, Zifu and Li, Shengxi and Liu, Henan and Xu, Mai and Zhu, Ce},
  journal={IEEE Signal Processing Letters},
  year={2025},
  publisher={IEEE}
}

@article{han2025generating,
  title={Generating inverse feature space for class imbalance in point cloud semantic segmentation},
  author={Han, Jiawei and Liu, Kaiqi and Li, Wei and Zhang, Feng and Xia, Xiang-Gen},
  journal={IEEE Transactions on Pattern Analysis and Machine Intelligence},
  year={2025},
  publisher={IEEE}
}

@article{han2024large,
  title={A large-scale network construction and lightweighting method for point cloud semantic segmentation},
  author={Han, Jiawei and Liu, Kaiqi and Li, Wei and Chen, Guangzhi and Wang, Wenguang and Zhang, Feng},
  journal={IEEE Transactions on Image Processing},
  volume={33},
  pages={2004--2017},
  year={2024},
  publisher={IEEE}
}

@ARTICLE{11165183,
  author={Shi, Jingjia and Zhi, Shuaifeng and Xu, Kai},
  journal={IEEE Transactions on Pattern Analysis and Machine Intelligence}, 
  title={PlaneRecTR++: Unified Query Learning for Joint 3D Planar Reconstruction and Pose Estimation}, 
  year={2025},
  volume={},
  number={},
  pages={1-18},
  doi={10.1109/TPAMI.2025.3610500}}

@article{liu2025causal,
  title={A causal adjustment module for debiasing scene graph generation},
  author={Liu, Li and Sun, Shuzhou and Zhi, Shuaifeng and Shi, Fan and Liu, Zhen and Heikkil{\"a}, Janne and Liu, Yongxiang},
  journal={IEEE Transactions on Pattern Analysis and Machine Intelligence},
  year={2025},
  publisher={IEEE}
}

\begin{IEEEbiography}[{\includegraphics[width=1in,height=1.25in,clip,keepaspectratio]{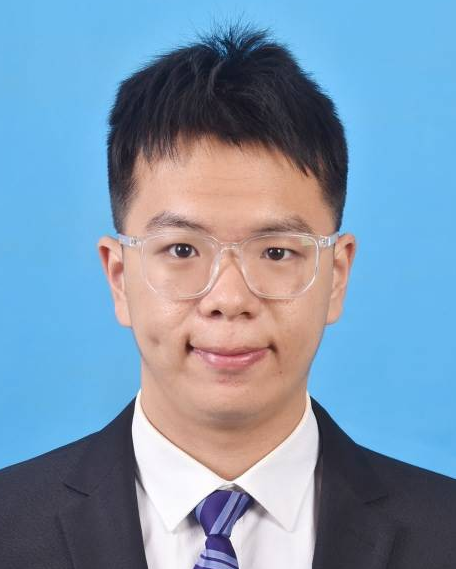}}]{Zifu Zhang}
(Student Member, IEEE) received the B.S. degree from the School of Electronics and Information Engineering, Beihang University, Beijing, China, in 2023. He is currently pursuing the M.S. degree with the School of Electronics and Information Engineering, Beihang University. His research interests mainly include learned image compression, image coding for machines, and deep generative models.
\end{IEEEbiography}

\begin{IEEEbiography}[{\includegraphics[width=1in,height=1.25in,clip,keepaspectratio]{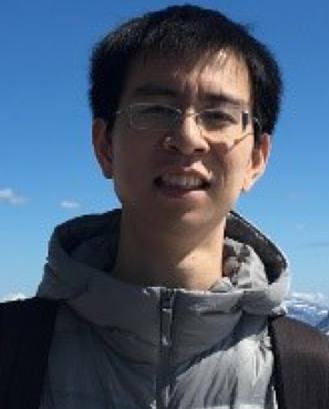}}]{Shengxi Li}
 (Member, IEEE) received the Ph.D. degree in electrical and electronic engineering from Imperial College London, London, U.K., in 2021. He is currently a Professor with the School of Electronic and Information Engineering, Beihang University, Beijing, China. His research interests include generative models, statistical signal processing, and machine learning. He was a recipient of the Young Investigator Award of International Neural Network Society.
\end{IEEEbiography}

\begin{IEEEbiography}[{\includegraphics[width=1in,height=1.25in,clip,keepaspectratio]{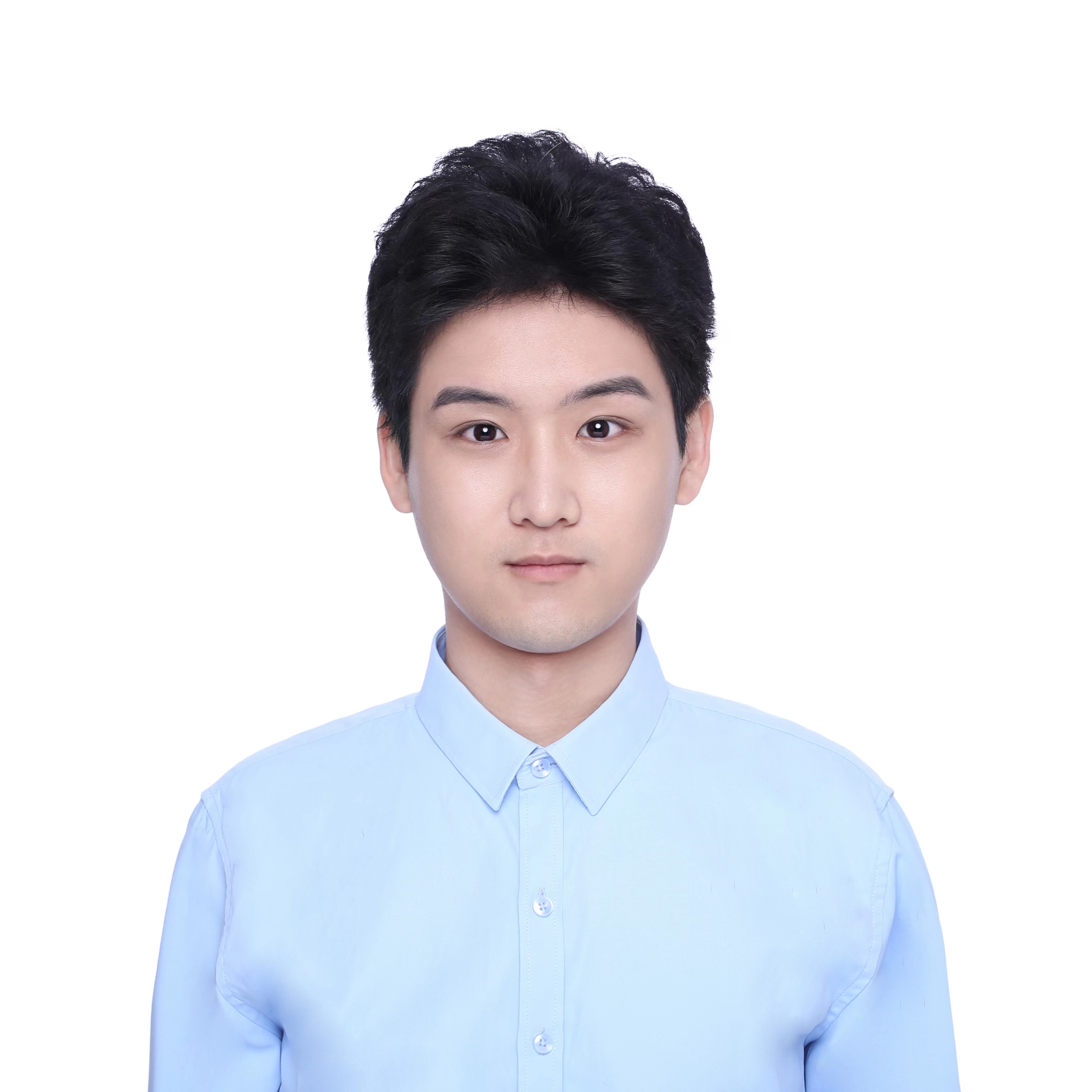}}]{Xiancheng Sun}
 (Graduate Student Member, IEEE) received the B.S. degree from the School of Elec
tronics and Information Engineering, Shen Yuan Honors College, Beihang University, Beijing, China, in 2023. He is currently pursuing the Ph.D. degree with the School of Electronics and Information Engineering, Beihang University. His research interests mainly include machine learning and generative models.
\end{IEEEbiography}

\begin{IEEEbiography}[{\includegraphics[width=1in,height=1.25in,clip,keepaspectratio]{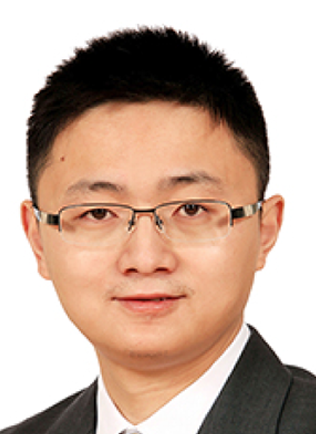}}]{Mai Xu}
  (Senior Member, IEEE) received the B.S. degree from Beihang University, Beijing, China, in 2003, the M.S. degree from Tsinghua University, Beijing, in 2006, and the Ph.D. degree from Imperial College London, London, U.K., in 2010. From 2010 to 2012, he was a Research Fellow with the Department of Electrical Engineering, Tsinghua University. Since 2013, he has been with Beihang University, where he was an Associate Professor and was promoted to a Full Professor in 2019. From 2014 to 2015, he was a Visiting Researcher with MSRA. He has authored or co-authored more than 200 technical papers in international journals and conference proceedings, such as IJCV, IEEE TRANSACTIONS ON PATTERN ANALYSIS AND MACHINE INTELLIGENCE, IEEE TRANSACTIONS ON IMAGE PROCESSING, CVPR, and ICCV. His main research interests include image processing and computer vision. He is an Elected Member of the Multimedia Signal Processing Technical Committee, IEEE Signal Processing Society. He was a recipient of the Best/Top Paper Awards of IEEE/ACM conferences, such as ACM MM. He was an Area Chair and a TPC Member of many conferences, such as ICME and AAAI. He served as an Associate Editor for IEEE TRANSACTIONS ON IMAGE PROCESSING and IEEE TRANSACTIONS ON MULTIMEDIA and a Lead Guest Editor for IEEE JOURNAL OFSELECTED TOPICS IN SIGNAL PROCESSING. He received Outstanding AE Awards in 2021 and 2022.
\end{IEEEbiography}

\begin{IEEEbiography}[{\includegraphics[width=1in,height=1.25in,clip,keepaspectratio]{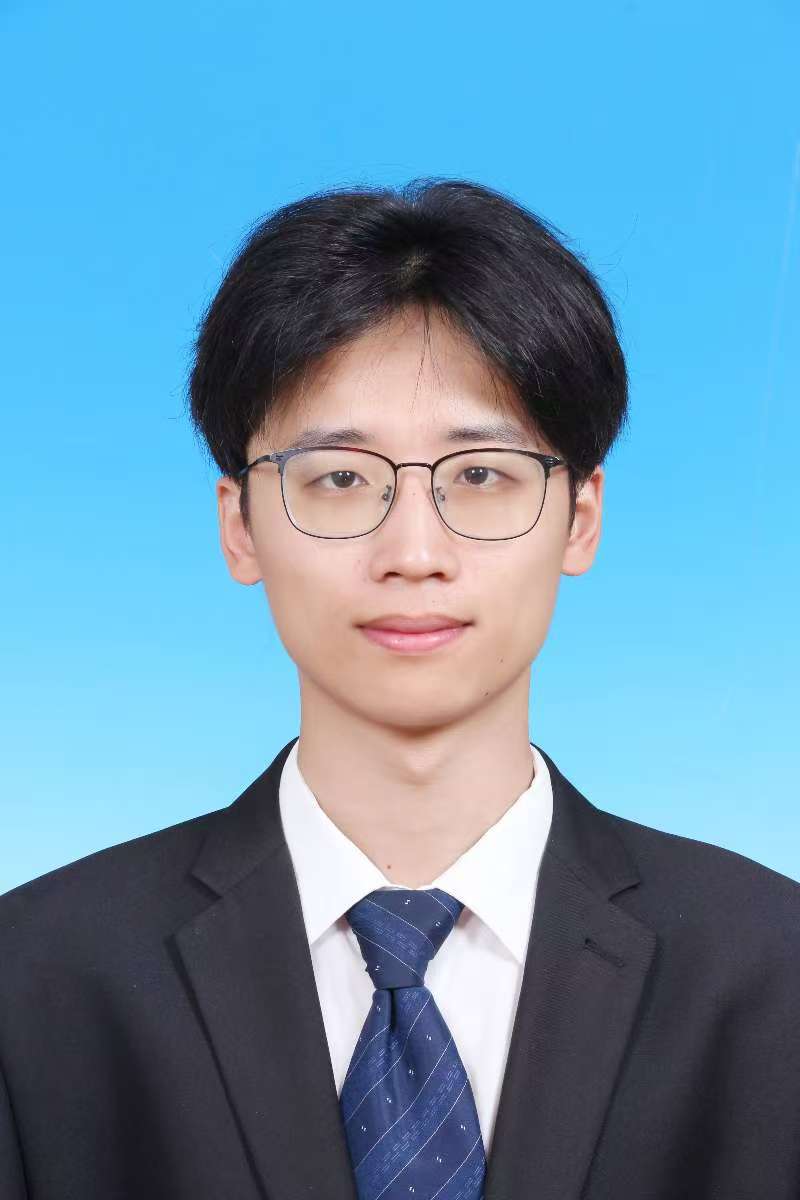}}]{Zhengyuan Liu}
is an M.S. candidate in the School of Electronics and Information Engineering at Beihang University, where he also completed his undergraduate studies in 2021. His research agenda is primarily in computer vision, with a particular interest in image compression technologies.
\end{IEEEbiography}

\begin{IEEEbiography}[{\includegraphics[width=1in,height=1.25in,clip,keepaspectratio]{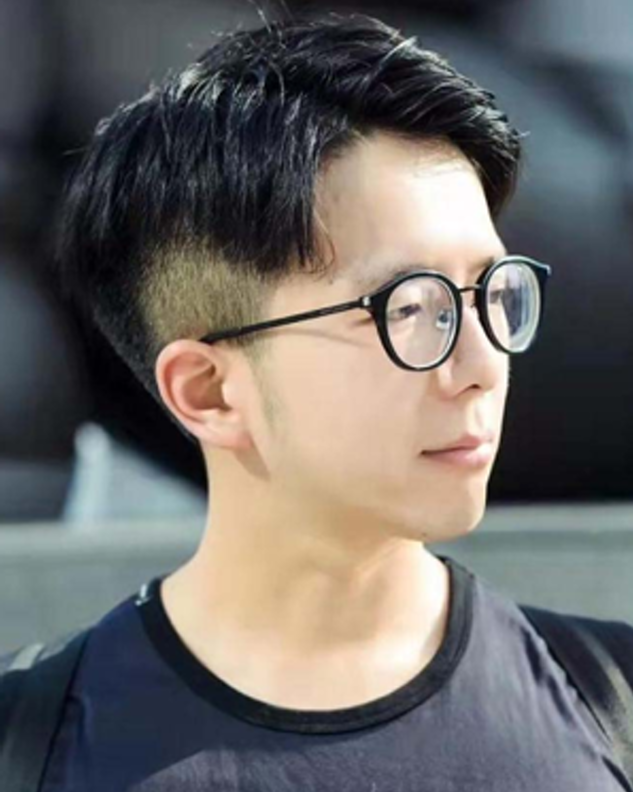}}]{Jingyuan Xia}
 (Member, IEEE) received the B.Sc. and M.Sc. degrees from the National University of Defense Technology (NUDT), Changsha, China, in 2014 and 2016, respectively, and the Ph.D. degree from Imperial College London (ICL), London, U.K., in 2020. He is currently an Associate Professor with the College of the Electronic Science, NUDT. His current research interests include low level image processing, nonconvex optimization, and machine learning for signal processing.
\end{IEEEbiography}



\end{document}